\documentclass[10pt,twocolumn,letterpaper,usenames,dvipsnames]{article}
\pdfoutput=1

\usepackage{iccv}
\usepackage{times}
\usepackage{epsfig}
\usepackage{graphicx}
\usepackage{amsmath,amssymb,bm}
\usepackage{multirow}
\usepackage{comment}

\usepackage[noadjust]{cite}

\usepackage[pagebackref=true,breaklinks=true,letterpaper=true,colorlinks,bookmarks=false]{hyperref}

\iccvfinalcopy 

\def\ie{\emph{i.e.}\xspace}
\def\eg{\emph{e.g.}\xspace}
\def\etal{\emph{et al.}\xspace}
\DeclareMathOperator*{\argmax}{argmax}
\newcommand{\com}[1]{{\textcolor{blue}{#1}}}

\newcommand{\cs}[1]{\textcolor{purple}{[\textbf{CS}: #1]}}

\newcommand{\myparagraph}[1]{\textbf{#1}\quad}
\newcommand{\eq}[1]{Eq.~(\ref{eq:#1})}
\newcommand{\fig}[1]{Figure~\ref{fig:#1}}

\ificcvfinal\fi
\begin{document}

\title{Objects2action: Classifying and localizing actions without any video example}

\author{Mihir Jain$^\star$ \\
\and
Jan C. van Gemert$^{\star\ddagger}$\\
\and
Thomas Mensink$^\star$\\
\and
Cees G. M. Snoek$^{\star\dagger}$ \\
}
\maketitle
\thispagestyle{empty}

\begin{abstract}
The goal of this paper is to recognize actions in video without the need for examples. Different from traditional zero-shot approaches we do not demand the design and specification of attribute classifiers and class-to-attribute mappings to allow for transfer from seen classes to unseen classes. Our key contribution is \emph{objects2action}, a semantic word embedding that is spanned by a skip-gram model of thousands of object categories. Action labels are assigned to an object encoding of unseen video based on a convex combination of action and object affinities.
Our semantic embedding has three main characteristics to accommodate for the specifics of actions. First, we propose a mechanism to exploit multiple-word descriptions of actions and objects. Second, we incorporate the automated selection of the most responsive objects per action. And finally, we demonstrate how to extend our zero-shot approach to the spatio-temporal localization of actions in video. Experiments on four action datasets demonstrate the potential of our approach.
\end{abstract}

\section{Introduction}
We aim for the recognition of actions such as \emph{blow dry hair} and \emph{swing baseball} in video without the need for examples. The common computer vision tactic in such a challenging setting is to predict the \emph{zero-shot} test classes from disjunct train classes based on a (predefined) mutual relationship using class-to-attribute mappings~\cite{Farhadi_cvpr09, lampert09cvpr, rohrbach2011evaluating, parikh2011relative, akata2013label}. Drawbacks of such approaches in the context of action recognition~\cite{JLiu_cvpr11} are that attributes like `torso twist' and `look-down' are difficult to define and cumbersome to annotate. Moreover, current zero-shot approaches, be it for image categories or actions, assume that a large, and labeled, set of (action) train classes is available a priori to guide the knowledge transfer, but today's action recognition practice is limited to at most hundred classes~\cite{ucf101,THUMOS14,hmdb51,Rodriguez:cvpr08}. Different from existing work, we propose zero-shot learning for action classification that does not require tailored definitions and annotation of action attributes, and not a single video or action annotation as prior knowledge.

\begin{figure}[t]
	\centering
	\includegraphics[width=\linewidth]{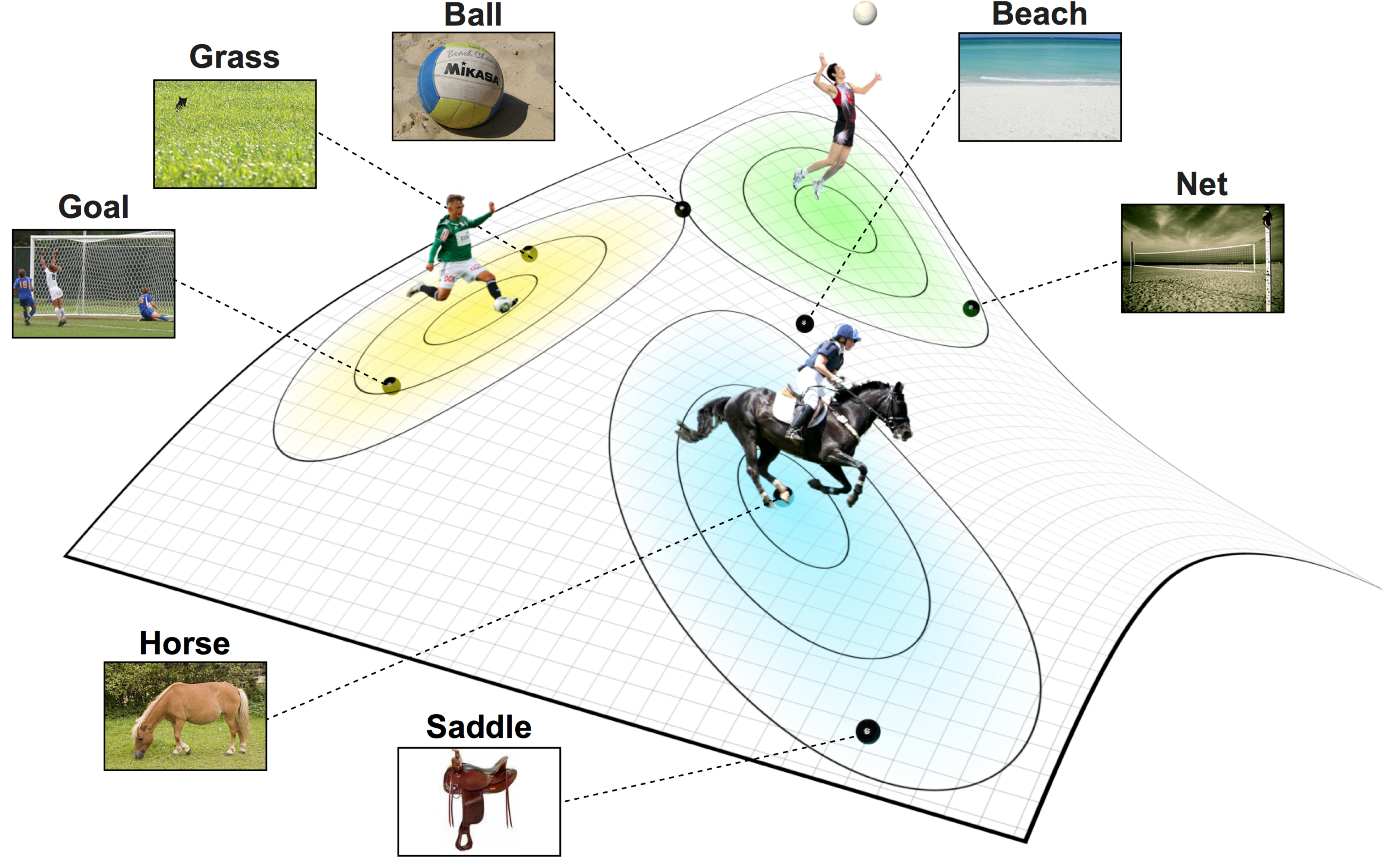}
	\caption{We propose \emph{objects2action}, a semantic embedding to classify actions, such as \emph{playing football}, \emph{playing volleyball}, and \emph{horse-riding}, in videos without using any video data or action annotations as prior knowledge. Instead it relies on commonly available textual descriptions, images and annotations of objects.}
	\label{fig:figure1}
\end{figure}

We are inspired by recent progress in \emph{supervised} video recognition, where several works successfully demonstrated the benefit of representations derived from deep convolutional neural networks for recognition of actions~\cite{karpathy2014, simonyan2014, Jain_15kObjAct} and events~\cite{SnoekTRECVID13, xu2015}. As these nets are typically pre-trained on images and object annotations from ImageNet~\cite{imagenet_cvpr09}, and consequently their final layer represent object category scores, these works reveal that object scores are well-suited for video recognition. Moreover, since these objects have a lingual correspondence derived from nouns in WordNet, they are a natural fit for semantic word embeddings~\cite{mikolov2013efficient_ICLR, socher2013zero, Conse_ICLR14, frome2013devise, elhoseiny:ICCV13}. 
As prior knowledge for our zero-shot action recognition we consider a semantic word embedding spanned by a large number of object class labels and their images from ImageNet, see Figure~\ref{fig:figure1}.

Our key contribution is \emph{objects2action}, a semantic embedding to classify actions in videos without using any video data or action annotations as prior knowledge. Instead it relies on commonly available object annotations, images and textual descriptions. Our semantic embedding has three main characteristics to accommodate for the specifics of actions. First, we propose a mechanism to exploit multiple-word descriptions of actions and ImageNet objects. Second, we incorporate the automated selection of the most responsive objects per action. And finally, we demonstrate our zero-shot approach to action classification and spatio-temporal localization of actions.

Before going into detail, we will first connect our approach to related work on action recognition and zero-shot recognition.

\section{Related work}

\subsection{Action Recognition}
The action classification literature offers a mature repertoire of elegant and reliable methods with good accuracy. Many methods include sampling spatio-temporal descriptors~\cite{evertsTIP14evaluation,wang:imptraj13}, aggregating the descriptors in a global video representation, such as versions of VLAD~\cite{Jain:wflow,shVlad_ecc14} or Fisher Vectors~\cite{pengECCV14stackedFisher} followed by supervised classification with an SVM. Inspired by the success of deep convolutional neural networks in image classification~\cite{Krizhevsky_imagenetclassification}, several recent works have demonstrated the potential of learned video representations for action and event recognition~\cite{karpathy2014, simonyan2014, Jain_15kObjAct, SnoekTRECVID13, xu2015}. All these deep representations are learned from thousands of object annotations~\cite{imagenet_cvpr09}, and consequently, their final output layer corresponds to object category responses indicating the promise of objects for action classification. We also use a deep convolutional neural network to represent our images and video as object category responses, but we do not use any action annotations nor any training videos.

Action classification techniques have recently been extended to action localization~\cite{Cao:cvpr10,tian_iccv11,Jain:tubelets,oneataECCV14spatemprops,gemertBMVC15apt} where in addition to the class, the location of the action in the video is detected. To handle the huge search space that comes with such precise localization, methods to efficiently sample action proposals~\cite{Jain:tubelets,oneataECCV14spatemprops,gemertBMVC15apt} are combined with the encodings and labeled examples used in action classification. In contrast, we focus on the zero-shot case where there is no labeled video data available for classification nor for localization. We are not aware of any other work on zero-shot action localization.

\subsection{Zero-Shot Recognition} 
The paradigm of zero-shot recognition became popular with the seminal paper of Lampert \etal~\cite{lampert09cvpr}.
The idea is that images can be represented by a vector of classification scores from a set of \emph{known classes}, and a semantic link can be created from the known class to a \emph{novel class}. Existing zero-shot learning methods can be grouped based on the different ways of building these semantic links.

A semantic link is commonly obtained by a human provided class-to-attribute mapping~\cite{akata2013label,Farhadi_cvpr09,lampert09cvpr,parikh2011relative}, where for each unseen class a description is given in terms of a set of attributes. Attributes should allow to tell classes apart, but should not be class specific, which makes finding good attributes and designing class-to-attribute mappings a non-trivial task. To overcome the need for human selection, at least partially, Rohrbach \etal evaluate external sources for defining the class-to-attribute mappings~\cite{rohrbach2011evaluating}. Typically, attributes are domain specific, \eg class and scene properties~\cite{lampert09cvpr} or general visual concepts~\cite{li14ijcv} learned from images, or action classes~\cite{actionbank_cvpr12} and action attributes~\cite{JLiu_cvpr11} learned from videos. 
{In our paper we exploit a diverse vocabulary of object classes for grounding unseen classes. Such a setup has successfully been used for action classification when action labels are available~\cite{Jain_15kObjAct}. In contrast, we have a zero-shot setting and do not use any action nor video annotations.}

{Zero-shot video event recognition as evaluated in TRECVID~\cite{over2013trecvid} offers meta-data in the form of a an event kit containing the event name, a definition, and a precise description in terms of salient concepts. Such meta-data can cleverly be used for a class-to-attribute mapping based on multi-modal concepts~\cite{habibianACM14zeroEventCompositeConcept,wuCVPR14zeroEventMultiModal}, seed a sequence of multimodal pseudo relevance feedback \cite{jiangICMR14zeroEventPseudoRel}, or select relevant tags from Flickr~\cite{chenICMR14eventWeakTag}. In contrast to these works, we do not assume any availability of additional meta-data and only rely on the action name.}

To generalize zero-shot classification beyond attribute-to-class mappings, Mensink \etal~\cite{mensink2014costa} explored various metrics to measure co-occurrence of visual concepts for establishing a semantic link between labels, and Froome \etal~\cite{frome2013devise} and Norouzi \etal~\cite{Conse_ICLR14} exploit semantic word embeddings for this link. We opt for the latter direction, and also use a semantic word embedding~\cite{leICML14paragraphVector,mikolov2013efficient_ICLR,mikolov2013distributed} since this is the most flexible solution, and allows for exploiting object and action descriptions containing multiple words, such as the WordNet synonyms and the subject, verb and object triplets to describe actions~\cite{guadarrama:ICCV13,sun2014semantic} used in this paper.

\section{Objects2action} \label{sec:method}

\begin{figure}[t]
	\centering
	\includegraphics[width=\linewidth]{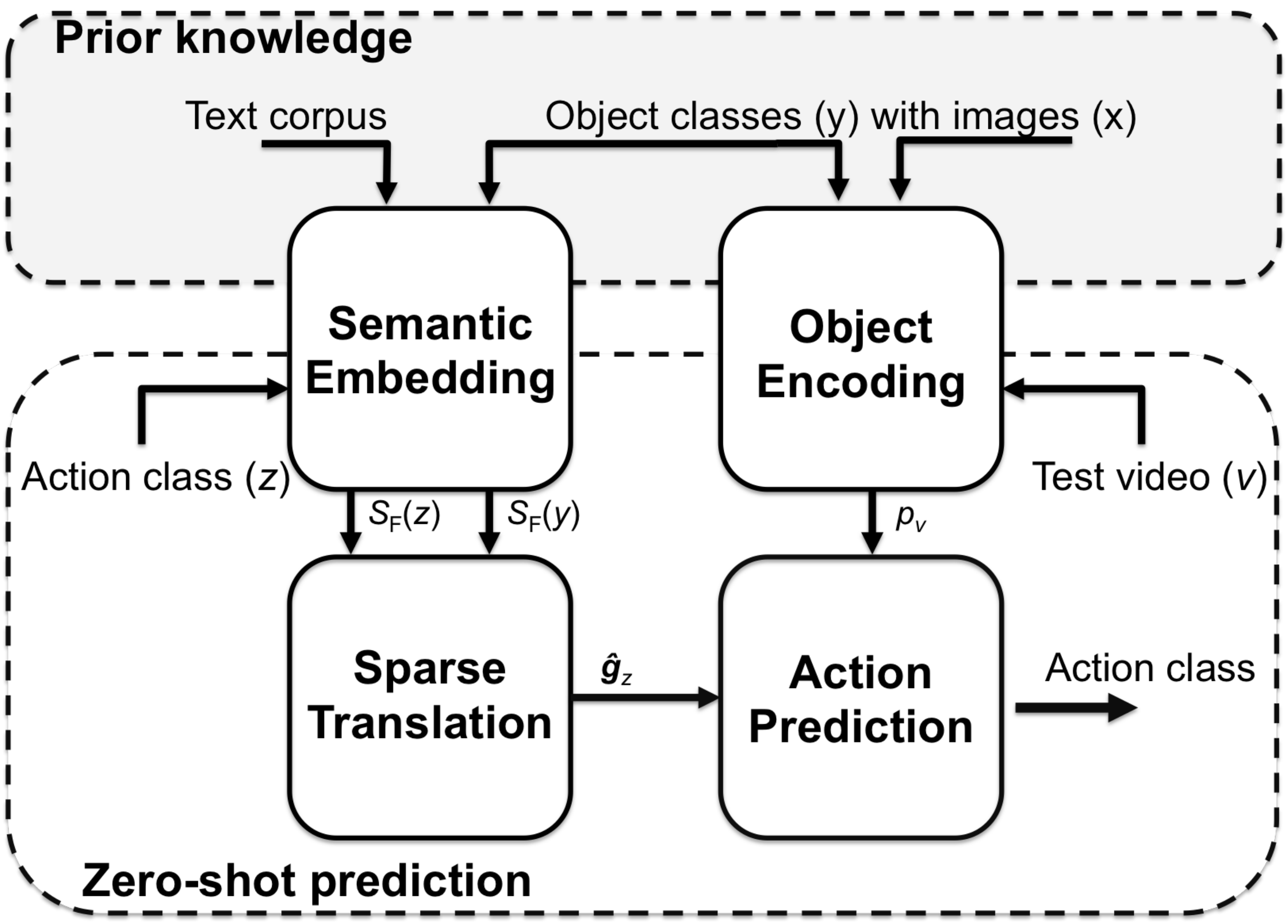}
	\caption{Dataflow in objects2action. Intermediate processes, data and corresponding symbols are specified in Section \ref{sec:method}. Sparse tranlsation is only shown for action to object affinity. Note that we do not require any action class labeled visual examples nor any video examples as prior knowledge.}
	\label{fig:pipeline}
\end{figure}

In zero-shot classification the train classes $\mathcal Y$ are different from the set of \emph{zero-shot} test classes $\mathcal Z$, such that $\mathcal Y \cap \mathcal Z = \emptyset$. For training samples $\mathcal X$, a labeled dataset $\mathcal D \equiv \{\mathcal X, \mathcal Y\}$ is available, and the objective is to classify a test sample as belonging to one of the test classes $\mathcal Z$.
Usually, test samples $v$ are represented in terms of classification scores for all train classes $p_{vy} \, \forall y \in \mathcal{Y}$, and an affinity score $g_{yz}$ is defined to relate these train classes to the test classes.
Then the zero-shot prediction could be understood as a convex combination of known classifiers~\cite{akata2013label,mensink2014costa,Conse_ICLR14}:
\begin{equation}
 \mathcal C(v) = \argmax_z  \sum_y \; p_{vy} \, g_{yz}.
\end{equation}

Often there is a clear relation between training classes $\mathcal Y$ and test classes $\mathcal Z$, for example based on class-to-attribute relations~\cite{akata2013label,lampert09cvpr} or all being nouns from the ImageNet/Wordnet hierarchy~\cite{frome2013devise,Conse_ICLR14}. 
It is unclear, however, how to proceed when train classes and test classes are semantically disjoint.

Our setup, see \fig{pipeline}, differs in two aspects to the standard zero-shot classification pipeline: i) 
our zero-shot test examples are videos $\mathcal V$ to be classified in actions $\mathcal Z$, while we have a train set $\mathcal D$ with images $\mathcal X$ labeled with objects $\mathcal Y$ derived from ImageNet~\cite{imagenet_cvpr09}. Therefore, we aim to \emph{transfer} from the domain of images $\mathcal X$ to the domain of videos $\mathcal{V}$, and  ii)  we aim to \emph{translate} objects semantics $\mathcal Y$ to the semantics of actions $\mathcal Z$.

\myparagraph{Object encoding}
We encode a test video $v$ by the classification scores to the $m = |Y|$ objects classes from the train set:
\begin{equation}
    \bm p_v = [p(y_1|v), \ldots, p(y_m|v)]^T
\end{equation}
where the probability of an object class is given by a deep convolutional neural network trained from ImageNet~\cite{Krizhevsky_imagenetclassification}, as recently became popular in the video recognition literature \cite{karpathy2014,simonyan2014,Jain_15kObjAct,SnoekTRECVID13,xu2015}. For a video $v$ the probability $p(y|v)$ is computed by averaging over the frame probabilities, where every $10^{th}$ frame is sampled. We exploit the semantics of in total 15,293 ImageNet object categories for which more than 100 examples are available. 

We define the affinity between an object class $y$ and action class $z$ as:
\begin{equation}
    g_{yz} = s(y)^T s(z),
    \label{eq:test_rep}
\end{equation}
where $s(\cdot)$ is a semantic embedding of any class $\mathcal Z \cup \mathcal Y$, and we use $\bm g_z = [s(y_1) \ldots s(y_m)]^T s(z)$ to represent the translation of action $z$ in terms of objects $\mathcal{Y}$.
The semantic embedding function $s$ is further detailed below.

\subsection{Semantic embedding via Gaussian mixtures}
\label{sec:semEmbed}
The objective for a semantic embedding is to find a $d$-dimensional space, in which the distance between an object $s(y)$ and an action $s(z)$ is small, if and only if their classes $y$ and $z$ are found in similar (textual) context.
For this we employ the skip-gram model of word2vec~\cite{mikolov2013efficient_ICLR,mikolov2013distributed} as semantic embedding function, which results in a look-up table for each word, corresponding to a $d$-dimensional vector. 

Semantic word embeddings have been used for zero-shot object classification~\cite{Conse_ICLR14}, but in our setting the key differences are i) that train and test classes come from different domains: objects in the train set and actions in the test set; and ii) both the objects and actions are described with a small description instead of a single word. In this section we describe two embedding techniques to exploit these multi-word descriptions to bridge the semantic gap between objects and actions. 

\myparagraph{Average Word Vectors (\emph{AWV})}
The first method to exploit multiple words is take the average vector of the embedded words~\cite{milajevs2014evaluating}.
The embedding $s(c)$ of a multi-words description $c$ is given by:
\begin{equation}
    s_{\textrm{A}}(c) = \frac{1}{|w|} \sum_{w \in c} s(w).
\end{equation}
This model combines words to form a single average word, as represented with a vector inside the word embedding. While effective, this cannot model any semantic relations that may exist between words. For example, the relations for the word \emph{stroke}, in the sequence \emph{stroke, swimming, water} is completely different than the word relations in the sequence \emph{stroke, brain, ambulance}.

\myparagraph{Fisher Word Vectors (\emph{FWV})}
To describe the precise meaning of distinct words we propose to aggregate the word embeddings using Fisher Vectors~\cite{sanchez13ijcv}.
While these were originally designed for aggregating local image descriptors~\cite{sanchez13ijcv}, they can be used for aggregating words as long as the \emph{discrete} words are transformed into a \emph{continuous} space~\cite{clinchant13ictir}.
In contrast to~\cite{clinchant13ictir}, where LSI is used to embed words into a continuous space, we employ the word embedding vectors of the skip-gram model. These vectors for each word are then analogous to local image descriptors and a class description is analogous to an image.

The advantage of the FWV model is that it uses an underlying generative model over the words. This generative model is modeling semantic topics within the word embedding. Where AWV models a single word, the FWV models a distribution over words. The \emph{stroke} example could for example be assigned to two clear, distinct topics \emph{infarct} and \emph{swimming}. This word sense disambiguation leads to a more precise semantic grounding at the topic-level, as opposed to single word-level.

In the Fisher Vector, a document (\ie a set of words) is described as the gradient of the log-likelihood of these observations on an underlying probabilistic model. Following~\cite{sanchez13ijcv} we use a diagonal Gaussian Mixture Model with $k$ components as probabilistic model, which we learn on approximately $45K$ word embedding vectors from the 15K object classes in ImageNet. 

The Fisher Vectors with respect to the mean $\mu_k$ and variance $\sigma_k$ of mixture component $k$ are given by:
\begin{align}
    \mathcal{G}_{\mu_k}^c &= \frac{1}{\sqrt{\, \pi_k}} \ \sum_{w \in c} \gamma_w(k) \left( \frac{s(w) - \mu_k}{\sigma_k} \right),\label{eq:fv1}\\
    \mathcal{G}_{\sigma_k}^c &= \frac{1}{\sqrt{2 \pi_k}} \sum_{w \in c} \gamma_w(k) \left( \frac{(s(w) - \mu_k)^2}{\sigma_k^2} - 1 \right),\label{eq:fv2}
\end{align}
where $\pi_k$ is the mixing weight, and $\gamma_w(k)$ denotes the responsibility of component $k$ and we use the closed-form approximation of the Fisher information matrix of~\cite{sanchez13ijcv}. 
The final Fisher Word Vector is the concatenation of the Fisher Vectors (\eq{fv1} and \eq{fv2}) for all components:
\begin{equation}
    s_{\textrm{F}}(c) = [\mathcal{G}_{\mu_1}^c, \mathcal{G}_{\sigma_1}^c,\ldots,\mathcal{G}_{\mu_k}^c,\mathcal{G}_{\sigma_k}^c]^T.
\end{equation}

\subsection{Sparse translations}
The action representation $\bm g_z$ represents the translation of the action to all objects in $\mathcal{Y}$.
However, not all train objects are likely to contribute to a clear description of a specific action class.
For example, consider the action class \emph{kayaking}, it makes sense to translate this action to object classes such as \emph{kayak}, \emph{water}, and \emph{sea}, with some related additional objects like \emph{surf-board}, \emph{raft}, and \emph{peddle}. 
Likewise a similarity value with, \eg, \emph{dog} or \emph{cat} is unlikely to be beneficial for a clear detection, since it introduces clutter. We consider two sparsity metrics that operate on the action classes or the test video.

\myparagraph{Action sparsity}
We propose to sparsify the representation $\bm g_z$ by selecting the $T_z$ most responsive object classes to a given action class $z$.
Formally, we redefine the action to object affinity as:
\begin{equation}
	 \hat{\bm g}_{z} = [g_{zy_1} \delta(y_1,T_z),\ldots,   g_{zy_m} \delta(y_m,T_z)]^T
\end{equation}
where $\delta(y_i,T_z)$ is an indicator function, returning 1 if class $y_i$ is among the top $T_z$ classes. 
In the same spirit, the objects could also have been selected based on their distance, considering only objects within an $\epsilon_z$ distance from $s(z)$. We opt for the selection of the top $T_z$ documents, since it is easier to define an a priori estimate of the value. Selecting $T_z$ objects for an action class $z$ means that we focus only on the object classes that are closer to the action classes in the semantic space. 

\myparagraph{Video sparsity}
Similarly, the video representation $\bm p_v$ is, by design, a dense vector, where each entry contains the (small) probability $p(y|v)$, of the presence of train class $y$ in the video $v$. 
We follow~\cite{Conse_ICLR14} and use only the top $T_v$ most prominent objects present in the video:
\begin{equation}
	\hat{\bm p}_v =  [p(y_1|v) \delta(y_1,T_v),\ldots,   p(y_m|v) \delta(y_m,T_v)]^T
\end{equation}
where $\delta(y_i,T_v)$ is an indicator function, returning 1 if class $y_i$ is among the top $T_v$ classes.
Increasing the sparsity, by considering only the top $T_v$ class predictions will reduce the effect of adding random noise by summing over a lot of unrelated classes with a low probability mass and is therefore likely to be beneficial for zero-shot classification. 

The optimal values for both $T_z$ and $T_v$ are likely to depend on the datasets, the semantic representation and the specific action description. Therefore they are considered as hyper-parameters of the model. Typically, we would expect that $T \ll m$, \eg, the 50 most responsive object classes will suffice for representing the video and finding the best action to object description.

\subsection{Zero-shot action localization}
Objects2action is easily extendable to zero-shot localization of actions by exploiting recent advances in sampling spatio-temporal tube-proposals from videos~\cite{Jain:tubelets,oneataECCV14spatemprops,gemertBMVC15apt}. Such proposals have shown to give a high localization recall with a modest set of proposals.

From a test video, a set $\mathcal{U}$ of spatio-temporal tubes are sampled~\cite{Jain:tubelets}. For each test video we simply select the maximum scoring tube proposal:
\begin{equation}
 \mathcal C(v) = \argmax_{z \in \mathcal{Z}, u \in \mathcal{U}_v} \sum_y \; p_{uy} \, g_{yz},
\end{equation}
where $u$ denotes a spatio-temporal tube proposal, and $\bm p_{uy}$ is the probability of the presence of object $y$ in region $u$.

For the spatio-temporal localization, a tube proposal contains a series of frames, each with a bounding-box indicating the spatial localization of the action. We feed just the pixels inside the bounding-box to the convolutional neural network to obtain the visual representation embedded in object labels. We will demonstrate the localization ability in the experiments.

\begin{figure*}[t]
\centering
  \includegraphics[width=0.49\linewidth]{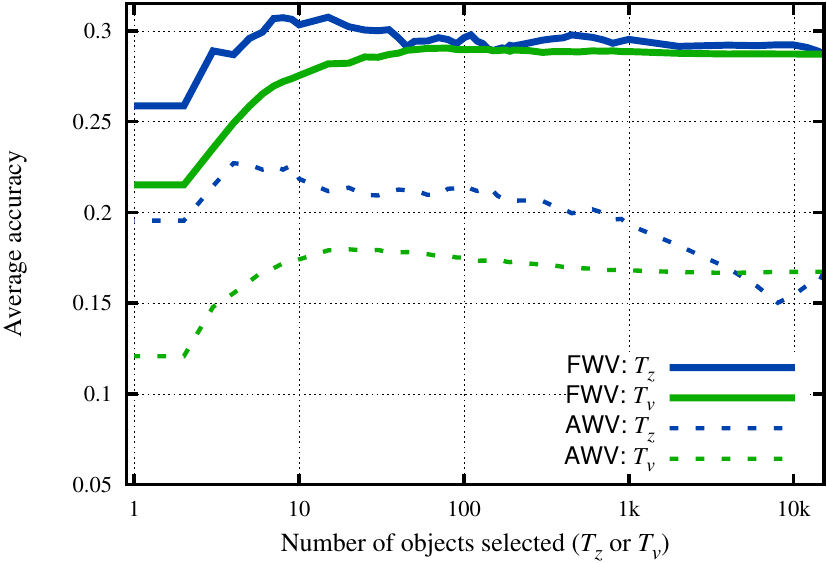}	
  \includegraphics[width=0.49\linewidth]{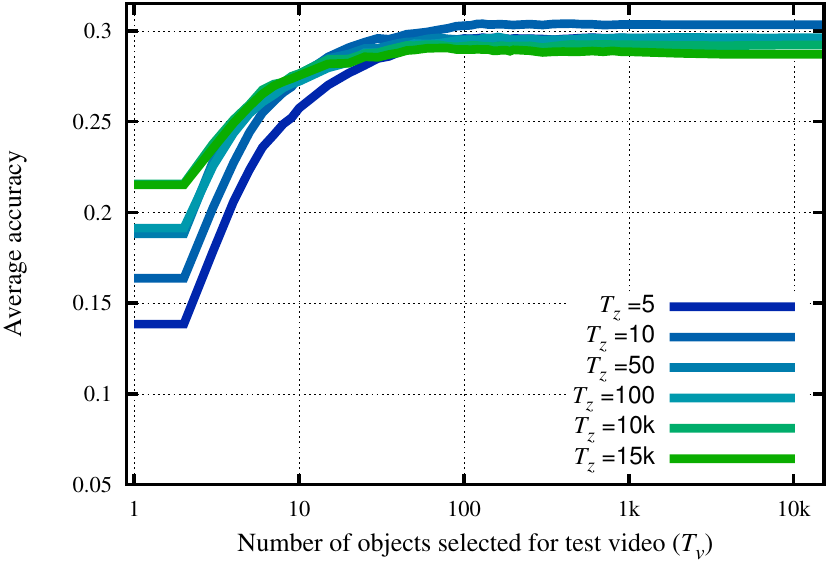}
  \caption{Impact of video $T_v$ and action $T_z$ sparsity parameters, individually (\emph{left}) and when combined (\emph{right}) on UCF101 dataset.}
\label{fig:impact_sparse}
\end{figure*}

\section{Experiments} \label{sec:exp}
In this section, we employ the proposed object2action model on four recent action classification datasets.
We first describe these datasets and the text corpus used.
Second, we analyze the impact of applying the Fisher Word Vector over the baseline of the Average Word Vector for computing the affinity between objects and actions, and we evaluate the action and video sparsity parameters. 
Third, we report zero-shot classification results on the four datasets, and we compare against  the traditional zero-shot setting where actions are used during training. 
Finally, we report performance of zero-shot spatio-temporal action localization.

\subsection{Prior knowledge and Datasets}
Our method is based on freely available resources which we use as prior knowledge for zero-shot action recognition. For the four action classification datasets datasets used we only use the \emph{test} set.

\myparagraph{Prior knowledge}
We use two types of prior knowledge.
First, we use deep convolutional neural network trained from ImageNet images with objects~\cite{Krizhevsky_imagenetclassification} as visual representation.
Second, for the semantic embedding we train the skip-gram model of word2vec on the metadata (title, descriptions, and tags) of the YFCC100M dataset~\cite{thomee15yfcc100m}, this dataset contains about 100M Flickr images. 
Preliminary experiments showed that using visual metadata results in better performance than training on Wikipedia or GoogleNews data. 
We attribute this to the more visual descriptions used in the YFC100M dataset, yielding a semantic embedding representing visual language and relations.

\myparagraph{UCF101~\cite{ucf101}} 
This dataset contains 13,320 videos of 101 action classes. 
It has realistic action videos collected from YouTube and has large variations in camera motion, object appearance/scale, viewpoint, cluttered background, illumination conditions, \etc. 
Evaluation is measured using average class accuracy, over the three provided test-splits with around 3,500 videos each.

\myparagraph{THUMOS14~\cite{THUMOS14}} 
This dataset has the same 101 action classes as in UCF101, but the videos are have a longer duration and are temporally unconstrained. 
We evaluate on the testset containing 1,574 videos, using mean average precision (mAP) as evaluation measure.

\myparagraph{HMDB51~\cite{hmdb51}} 
This dataset contains 51 action classes and 6,766 video clips extracted from various sources, ranging from YouTube to movies, and hence this dataset contains realistic actions.
Evaluation is measured using average class accuracy, over the three provided test-splits with each 30 videos per class (1,530 videos per split).

\myparagraph{UCF Sports~\cite{Rodriguez:cvpr08}} 
This dataset contains 150 videos of 10 action classes. 
The videos are from sports broadcasts capturing sport actions in dynamic and cluttered environments.
Bounding box annotations are provided and this dataset is often used for spatio-temporal action localization. 
For evaluation we use the test split provided by~\cite{tian_iccv11} and performance is measured by average class accuracy.

\begin{table}[t]
\centering
{\small
\begin{tabular}{| c | l | c | c | }
\hline
{\bf Embedding} & {\bf Sparsity}	&  {\bf Best }   	&   {\bf Accuracy at}   \\	
		& 			&  {\bf accuracy}	&   $T_z$=10, $T_v$=100 	\\ \hline	
		&	Video  		&    18.0\%		&       17.5\%	  	\\	
AWV		&	Action 		&    22.7\%		&       21.9\% 		\\		
		&	Combine		&    22.7\%		&	 21.6\%		\\	\hline	
		&	Video 	  	&    29.1\%         	&      29.0\%     	\\	
FWV		&	Action 		&    30.8\%         	&      30.3\%	 	\\		
		&	Combine    	&    30.8\%    		&    	30.3\%		\\  \hline	
\end{tabular}}
\caption{Evaluating AWV and FWV for object to class affinity, and comparing action and video sparsity on UCF101 dataset.}
\label{table:sparse_comb}
\end{table}

\subsection{Properties of Objects2action}
\label{sec:params}

\begin{table*}[t]
\centering
{\small
\renewcommand{\tabcolsep}{3pt}
\begin{tabular}{| c | l | c | c | c | c | }
\hline
{\bf Embedding}		& {\bf Sparsity}  &    {\bf UCF101}   &   {\bf HMDB51}	&  {\bf THUMOS14} & {\bf UCF Sports}    \\      \hline
\multirow{3}{*}{AWV}		& None		&    16.7\% 	        &      8.0\%        &    4.4\%        & 13.9\%       \\	 	
					& Video		&    17.5\%        	&      7.7\%        &   10.7\%        & 13.9\%       \\
					& Action		&    21.9\%        	&      9.9\%        &   19.9\%        & 25.6\%       \\	\hline						
\multirow{3}{*}{FWV}	   	& None		&    28.7\%  	    	&     14.2\%        &    25.9\%        & 23.1\%       \\			
                    			& Video		&    29.0\%         	&     14.5\%        &   27.8\%        & 23.1\%       \\
					& Action		&   {\bf 30.3\%}    	&   {\bf 15.6\%}    & {\bf 33.4\%}    & {\bf 26.4\%}  \\    \hline \hline
\multicolumn{2}{|c|}{{Supervised}} &   63.9\%    &  35.1\%   	& 56.3\%		&  60.7\%	\\	\hline		
\end{tabular}}
\caption{{Evaluating semantic embeddings, action and video spartsity: Average accuracies (mAP for THUMOS14) for the four datasets. Action sparsity and FWV both boost the performance consistently. Supervised upper-bound using object scores as representation.}}
\label{table:cls_comp}
\end{table*}

\myparagraph{Semantic embedding} 
We compare the AWV with the FWV as semantic embedding.
For the FWV, we did a run of preliminary experiments to find suitable parameters for the number of components (varying $k=\{1,2,4,8,16,32\}$), the partial derivatives used (weight, mean, and/or variance)  and whether to use PCA or not.
We found them all to perform rather similar in terms of classification accuracy. 
Considering a label has only a few words (1 to 4), we therefore fix $k=2$, apply PCA to reduce dimensionality by a factor of 2, and to use only the partial derivatives \wrt the mean (conforming the results in ~\cite{clinchant13ictir}).
Hence, the total dimensionality of FWV is $d$, equivalent to the dimensionality of AWV, which allows for a fair comparison.
The two embeddings are compared in Table~\ref{table:sparse_comb} and Figure~\ref{fig:impact_sparse} (\emph{left}), and  FWV clearly outperforms AWV in all the cases.

\myparagraph{Sparsity parameters} 
In Figure~\ref{fig:impact_sparse},  we evaluate the action sparsity and video sparsity parameters.
The left plot shows average accuracy versus $T_z$ and $T_v$. 
It is evident that action sparsity, \ie, selecting most responsive object classes for a given action class leads to a better 
performance than video sparsity. 
The video sparsity (green lines) is more stable throughout and achieves best results in the range of 10 to 100 objects. 
Action sparsity is a bit sinuous, nevertheless it always performs better, independent of the type of embedding. 
Action sparsity is at its best in the range of selecting the 5 to 30 most related object classes. 
For the remaining experiments, we fix these parameters as $T_z=10$ and $T_v=100$.

We also consider the case when we apply sparsity on both video and actions (see the right plot). 
Applying sparsity on both sides does not improve performance, it is equivalent to the best action sparsity setting, showing that selecting the most prominent objects per action suffice for zero-shot action classification.
Table~\ref{table:sparse_comb} summarise the accuracies for the best and fixed choices of $T_z$ and $T_v$.

\subsection{Zero-shot action classification}	 \label{sec:act_cls}
In this section we employ the obtained parameters of Object2action, from the previous section, for zero-shot action classification on the test splits of all four  datasets.
We evaluate the benefit of using the FWV over AWV, and the effect of using sparsity (video sparsity, action sparsity or no sparsity at all).
The results are provided in Table~\ref{table:cls_comp}. 
We observe that the FWV always outperforms AWV, and that it is always beneficial to apply sparsity, and action sparsity with FWV performs the best.
{We also provide the supervised upper-bounds using the same video representation of object classification scores in Table~\ref{table:cls_comp}. Here and for all the experiments, 
we power normalize ($\alpha = 0.5$) the video representations before applying $\ell_2$ normalization.}

\myparagraph{Comparison to few-shot supervised learning}
In this experiment we compare the zero-shot classifier against few-shot supervised learning, on the THUMOS14 dataset.
For this we consider two types of video representation. 
The first representations, uses the state-of-the-art motion representation of~\cite{wang:imptraj13}, by encoding robust MBH descriptors along the improved trajectories~\cite{wang:imptraj13} using Fisher Vectors.
We follow the standard parameter cycle, by applying PCA, using a GMM with $K=256$ Gaussians, employing power and $\ell_2$ normalization.
The second representation uses the object scores $\bm p_v$ of a video, here also we apply power and $\ell_2$ normalization. 
For both representations, we train one-vs-rest linear SVM classifiers and we average performance over 20 runs for every given number of train examples.

\begin{figure}[t]
\centering
  \includegraphics[width=0.9\linewidth]{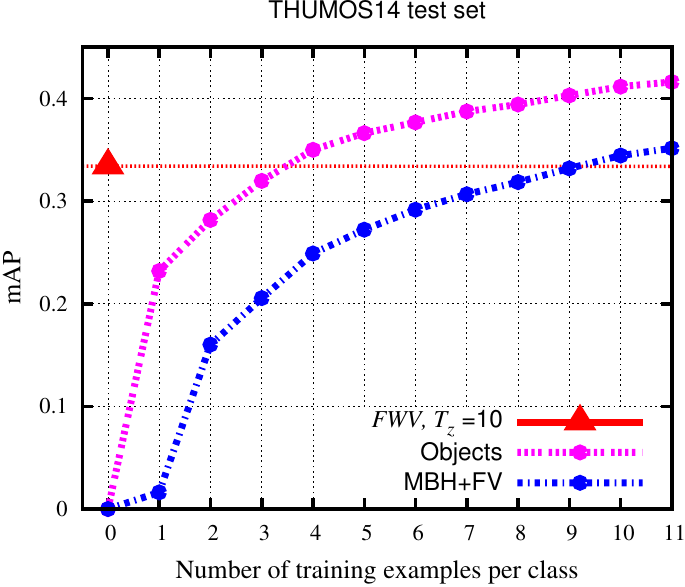}
  \caption{Our approach compared with the supervised classification with few examples per class: State-of-the-art object and motion representations respectively {require 4 and 10 examples per class} to catch up with our approach, which uses no example.}
\label{fig:thumos_cls}
\end{figure}

The results in mAP are shown in Figure~\ref{fig:thumos_cls}.
Interestingly, to perform better than our zero-shot classification, fully supervised classification setup requires 4 and 10 samples per class for object and motion representations respectively.

\myparagraph{Object transfer versus action transfer}
We now experiment with the more conventional setup for zero-shot learning, where we have training data for some action classes, disjoint from the set of test action classes. We keep half of the classes of a given dataset as train labels and the other half as our zero-shot classes. The action classifiers are learned from odd (or even) numbered classes and videos from the even (or odd) numbered classes are tested. 

{We evaluate two types of approaches for action transfer, \ie, when training classes are also actions. 
The first method uses the provided action attributes for zero-shot classification with direct attribute prediction~\cite{lampert09cvpr}. Since attributes are available only for 
UCF101, we experiment on this dataset. The train videos of the training classes are used to learn linear SVMs for the provided 115 attributes. 
The second method uses action labels embedded by FWV to compute affinity between train and test action labels. We use the same GMM with $k=2$ components learned on 
ImageNet object labels. Here linear SVMs are learned for the training action classes. The results are reported for UCF101 and HMDB51 datasets.
For both the above approaches for action transfer, we use MBH descriptors encoded by Fisher vectors for video representation.
The results are reported in Table~\ref{table:obj_act_comp}.}

{For comparison with our approach, the same setup of testing on odd or even numbered classes is repeated with the object labels. The training set is ImageNet objects, so no video example 
is used for training.}
Table~\ref{table:obj_act_comp} compares object transfer and action transfer for zero-shot classification. 
{Object transfer leads to much better learning compared to both the methods for action transfer.}
The main reason for the inferior performance using actions is that there are not enough action labels or action attributes to describe the test classes, whereas from 15k objects there is 
a good chance to find a few related object classes.

\begin{table}[t]
\centering
{\small
\renewcommand{\tabcolsep}{5pt}
\begin{tabular}{| c | l | l | c | c | }
\hline
{\bf Method}             &  {\bf Train}    &  {\bf Test} &  {\bf UCF101}      &  {\bf HMDB51} \\      \hline \hline
\multirow{ 2}{*}{Action attributes}        &     Even     &       Odd       	&     16.2\%         &  ---                \\
			                   &     Odd      &     Even  	  	&     14.6\%          &   ---                \\      \hline
\multirow{ 2}{*}{Action labels}   	 &     Even            &       Odd       &     16.1\%         & 12.4\%                \\
                 	 &     Odd      &       Even      &     14.4\%         & 13.4\%                \\      \hline \hline
\multirow{ 2}{*}{Objects2action}	&  \multirow{ 2}{*}{ImageNet}  &      Odd   	&     {\bf 37.3\%}       &  {\bf 15.0\%}       \\      
                    		&					&       Even      &     {\bf 38.9\%}     &  {\bf 24.5\%}            \\      \hline
\end{tabular}}
 \caption{{Object transfer versus action transfer in a conventional zero-shot set-up. Direct attribute prediction~\cite{lampert09cvpr} is used with action attributes, FWV is used to embed action labels, and in our objects2action.}}
\label{table:obj_act_comp}
\end{table}

{
\myparagraph{Zero-shot event retrieval}
We further demonstrate our method on the related problem of zero-shot event retrieval. We evaluate on the TRECVID13 MED~\cite{over2013trecvid} testset for EK0 task. 
There are 20 event classes and about 27,000 videos in the testset.
Instead of using the manually specified event kit containing the event name, a definition, and a precise description in terms of salient attributes, we only rely on the class label. In Table~\ref{table:med_comp}, we report mAP using event labels embedded by AWV and FWV. We also compare with the state-of-the-art approaches of Chen \etal~\cite{chenICMR14eventWeakTag} and Wu \etal~\cite{wuCVPR14zeroEventMultiModal} reporting their settings that are most similar to ours. They learn concept classifiers from images (from Flickr, Google) or YouTube video thumbnails, be it that they also use the complete event kit description. Using only the event labels, both of our semantic embeddings outperform these methods.} 

\begin{table}[t]
\centering
{\small
\begin{tabular}{| l | c |  c | }
\hline
\multicolumn{2}{|c|}{\bf Method}             						&    {\bf mAP}  \\      \hline 
\multicolumn{2}{|l|}{Wu \etal~\cite{wuCVPR14zeroEventMultiModal} (Google images)}             	&     1.21\%               \\      
\multicolumn{2}{|l|}{Chen \etal~\cite{chenICMR14eventWeakTag}    (Flickr images)}           	&     2.40\%               \\      
\multicolumn{2}{|l|}{Wu \etal~\cite{wuCVPR14zeroEventMultiModal} (YouTube thumbnails)}        	&     3.48\%               \\      \hline \hline
\multirow{ 2}{*}{Objects2action}   & 	AWV		         &     3.49\%                \\
			 &	FWV                      &     {\bf 4.21}\%               \\     
\hline
\end{tabular}}
\caption{{Zero-shot event retrieval on TRECVID13 MED testset: Comparison with the state-of-the-art methods having similar zero-shot setup as ours. Inspite of using only event labels and images, we outperform
methods that use event description and video thumbnails.}}
\label{table:med_comp}
\end{table}

\myparagraph{Free-text action search}
As a final illustration we show in Figure~\ref{fig:visualExamples} qualitative results from free-text querying of action videos from the THUMOS14 testset. We used the whole dataset for querying, and searched for actions that are not contained in the 101 classes of THUMOS14. Results show that free-text querying offers a tool to explore a large collection of videos. Results are best when the query is close to one or a few existing action classes, for example ``Dancing" retrieves results from ``salsa-spin" and other dancing clips. Our method fails for the query ``hit wicket", although it does find cricket matches. Zero shot action recognition through an object embedding unlocks free text querying without using any kind of expensive video annotations. 

\subsection{Zero-shot action localization}

\begin{figure}[t]
\centering
  \includegraphics[width=0.84\linewidth]{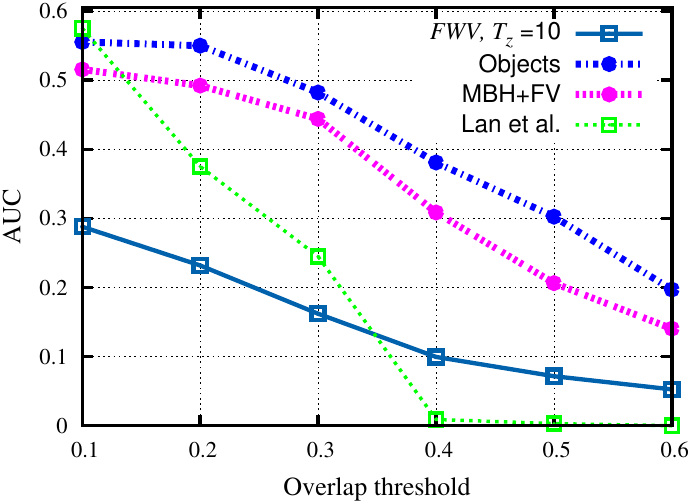}
  \caption{Action localization without video example on UCF Sports: AUCs for different overlap thresholds are shown for $T_z=10$ and also for the fully supervised setting with motion and object representations. The performance is promising considering no example videos are used.}
\label{fig:ST_loc}
\end{figure}


\begin{figure*}[t]
\begin{minipage}{1\linewidth}
\begin{minipage}{0.19\linewidth}
\centering  {\bf Fight in ring}
\includegraphics[width=0.98\linewidth]{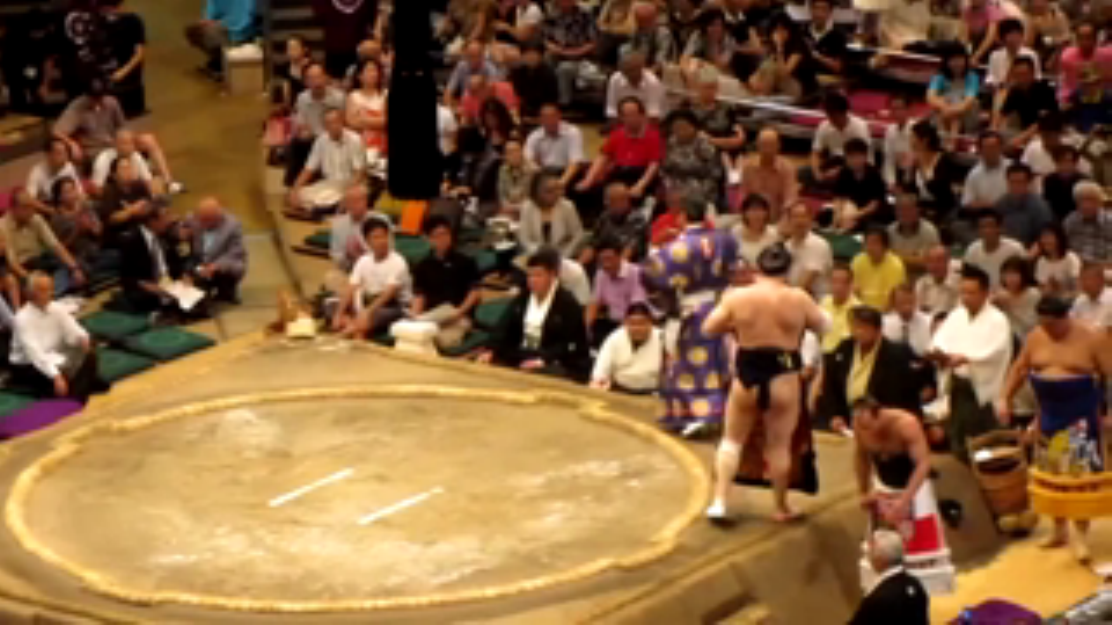}  \\
\includegraphics[width=0.98\linewidth]{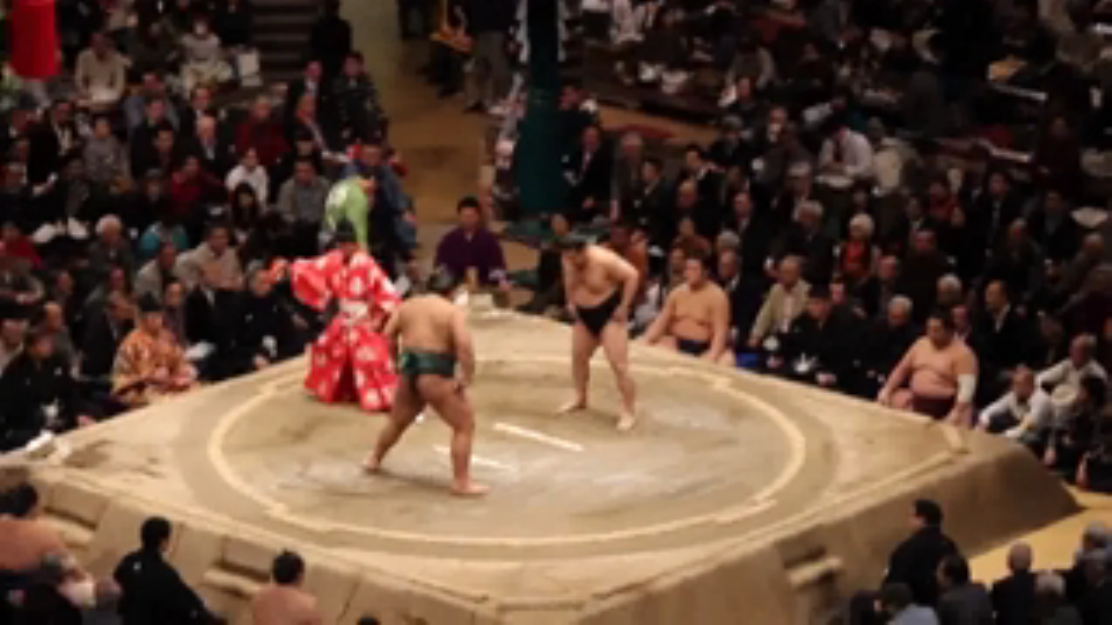}  \\
\includegraphics[width=0.98\linewidth]{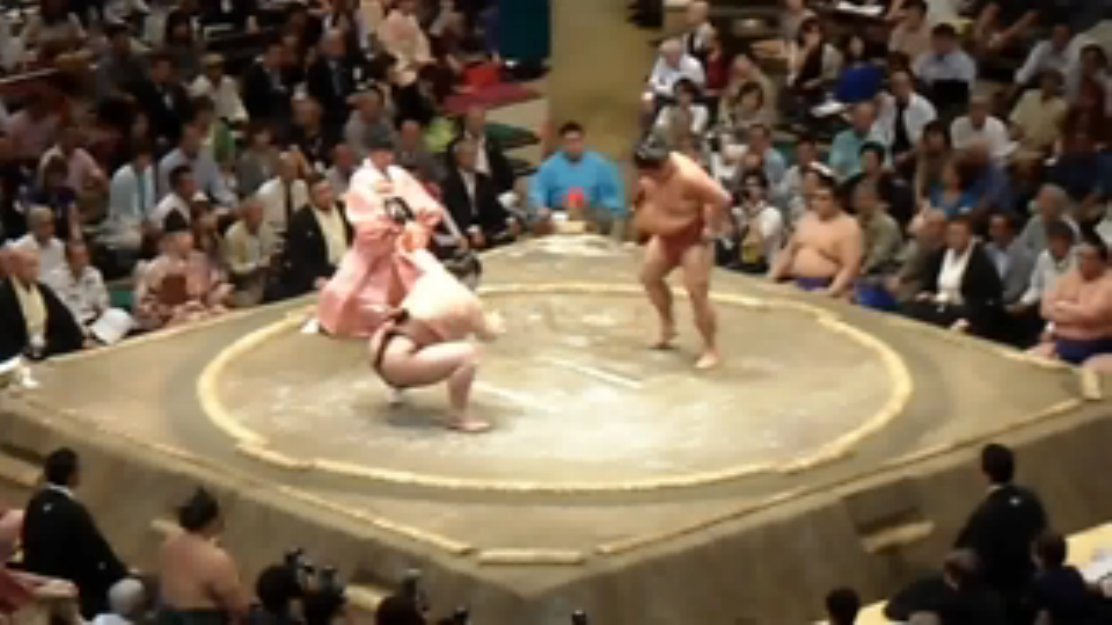}  \\
\includegraphics[width=0.98\linewidth]{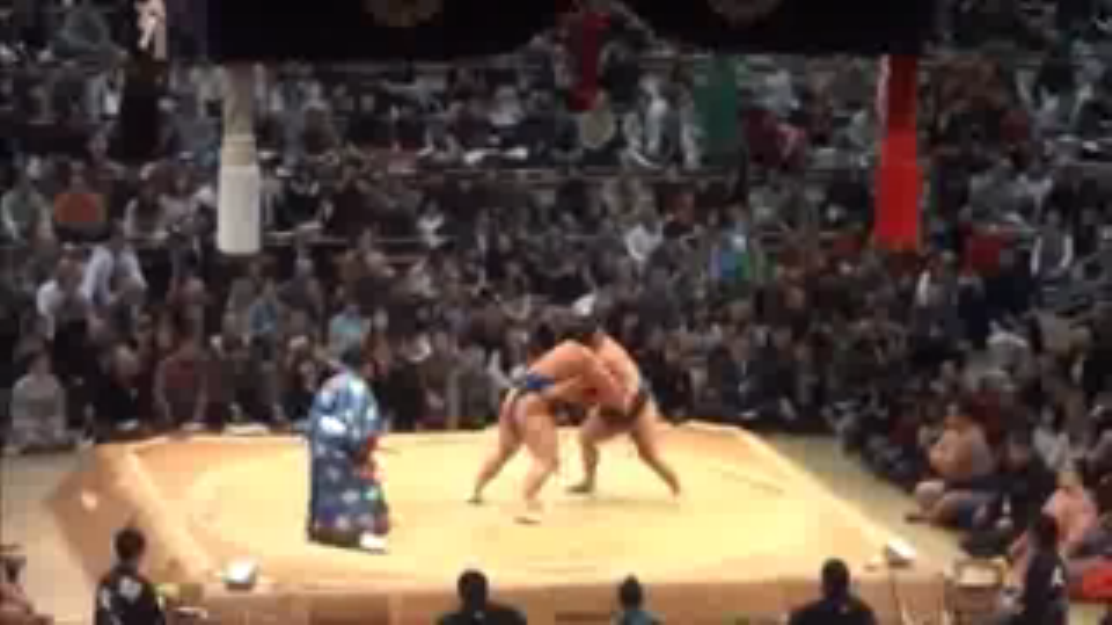}  \\
\includegraphics[width=0.98\linewidth]{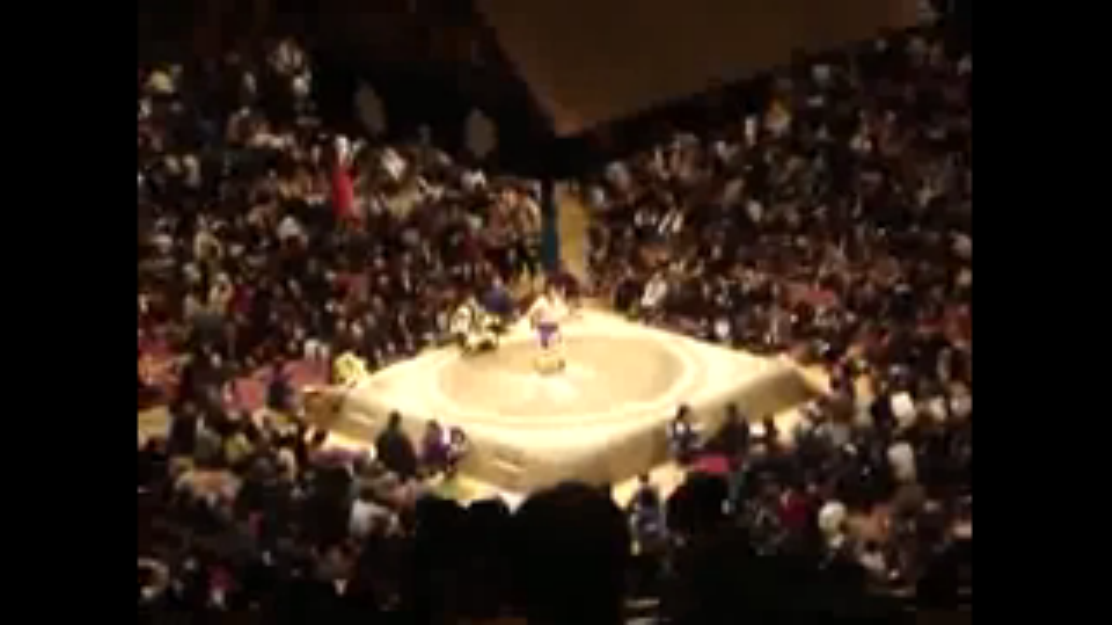}
\end{minipage}
\hfill
\begin{minipage}{0.19\linewidth}
\centering  {\bf Dancing}
\includegraphics[width=0.98\linewidth]{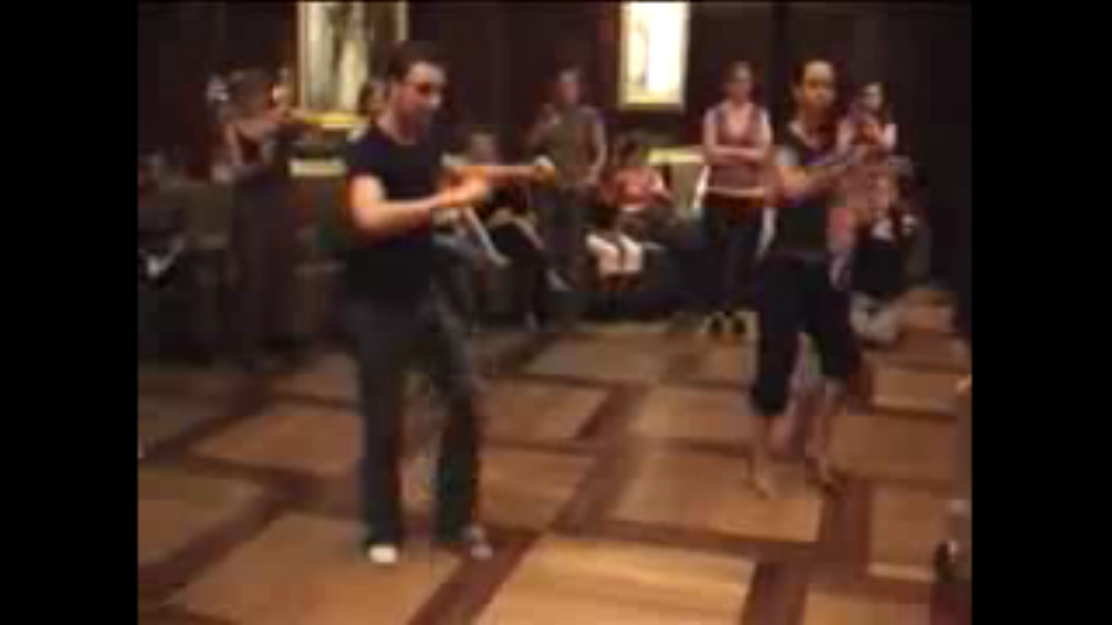}  \\
\includegraphics[width=0.98\linewidth]{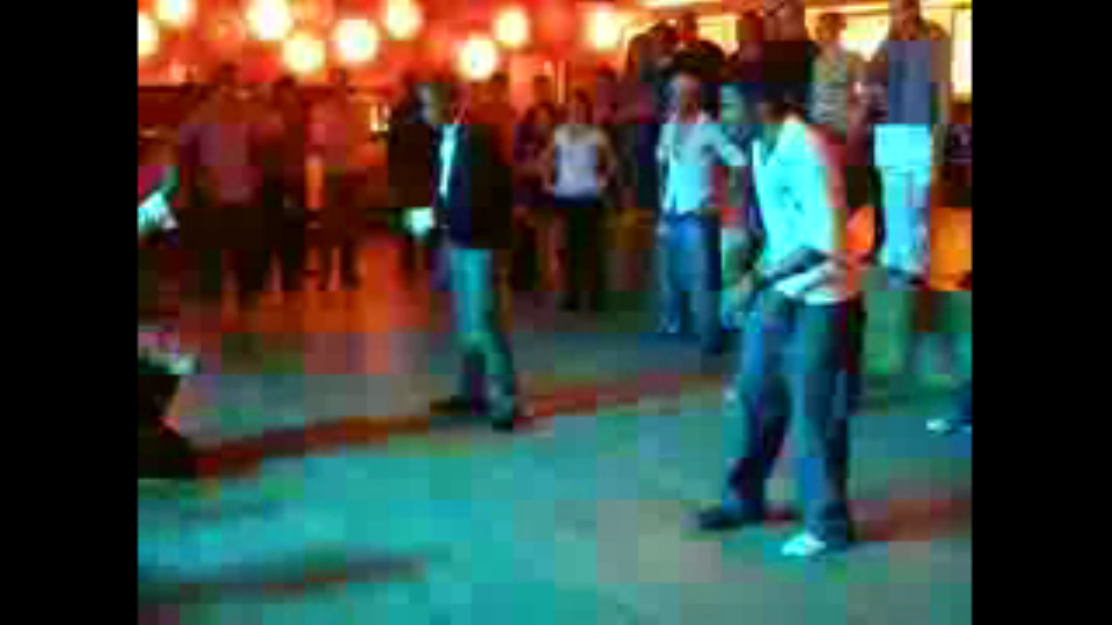}  \\
\includegraphics[width=0.98\linewidth]{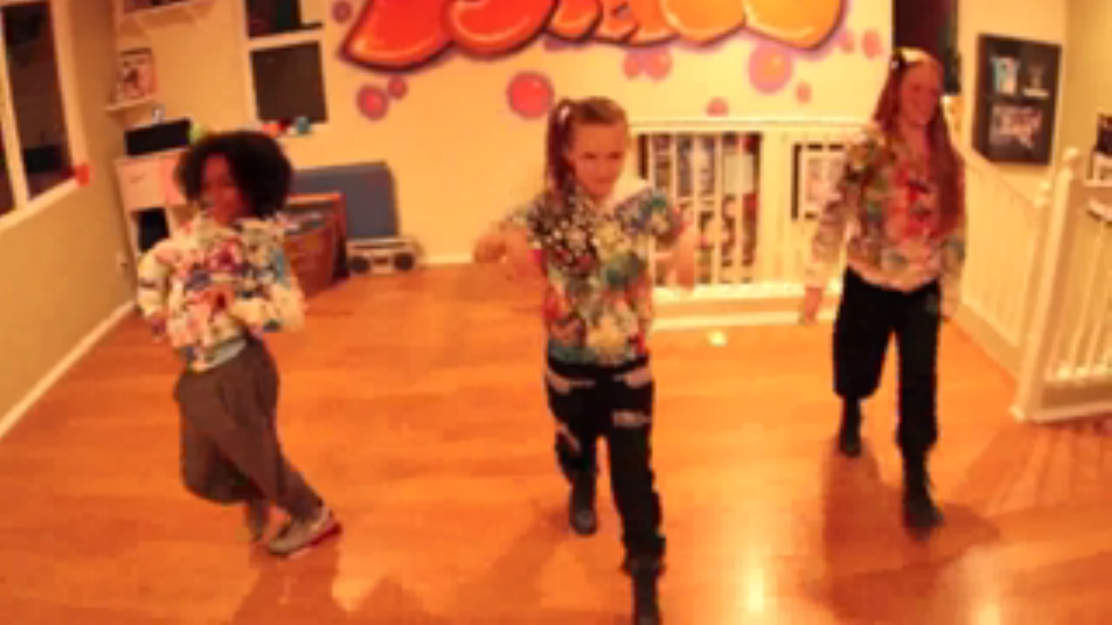}  \\
\includegraphics[width=0.98\linewidth]{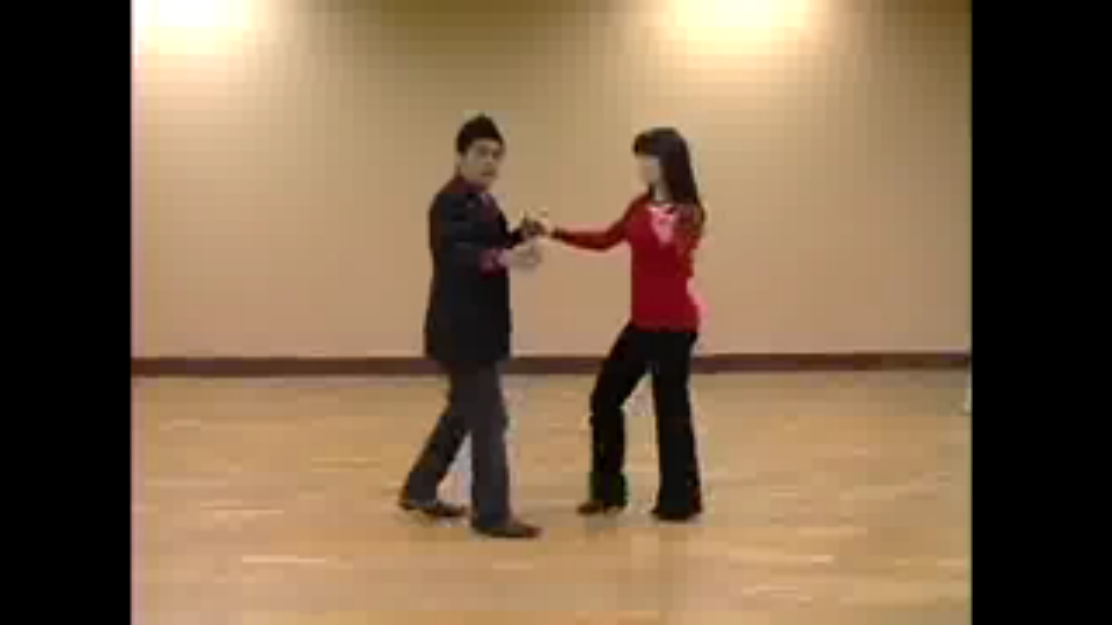}  \\
\includegraphics[width=0.98\linewidth]{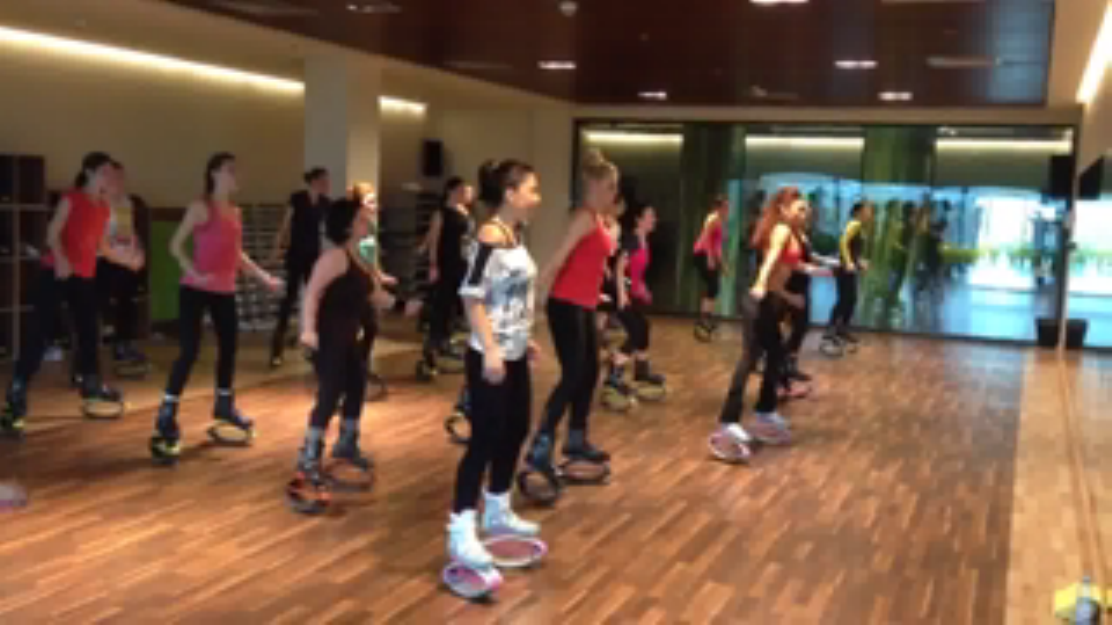}
\end{minipage}
\hfill
\begin{minipage}{0.19\linewidth}
\centering  {\bf Martial arts}
\includegraphics[width=0.98\linewidth]{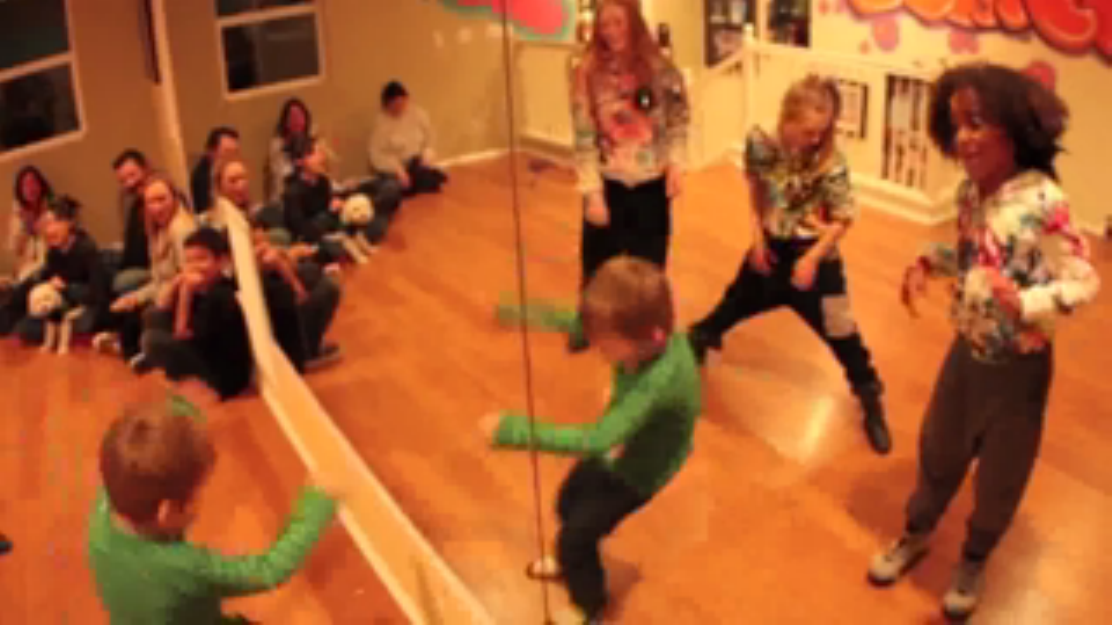}  \\
\includegraphics[width=0.98\linewidth]{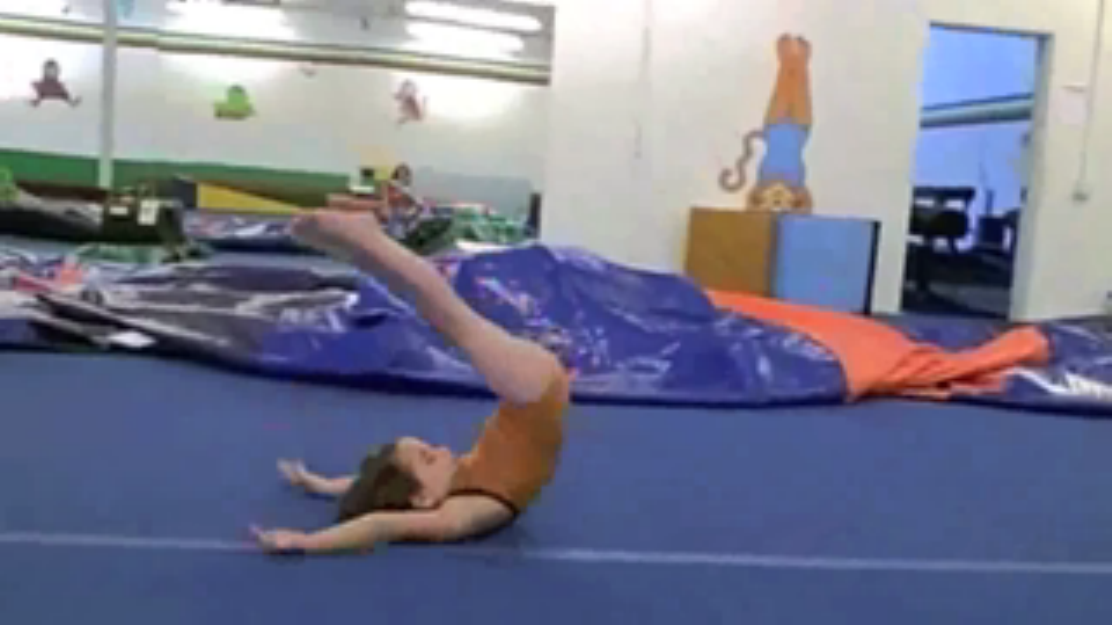}  \\
\includegraphics[width=0.98\linewidth]{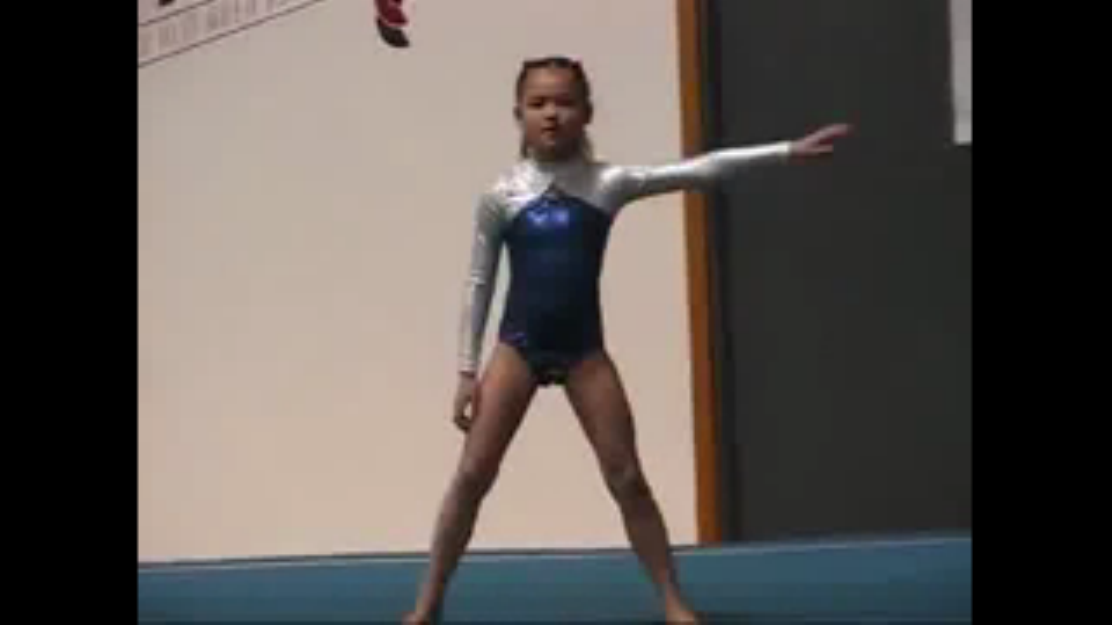}  \\
\includegraphics[width=0.98\linewidth]{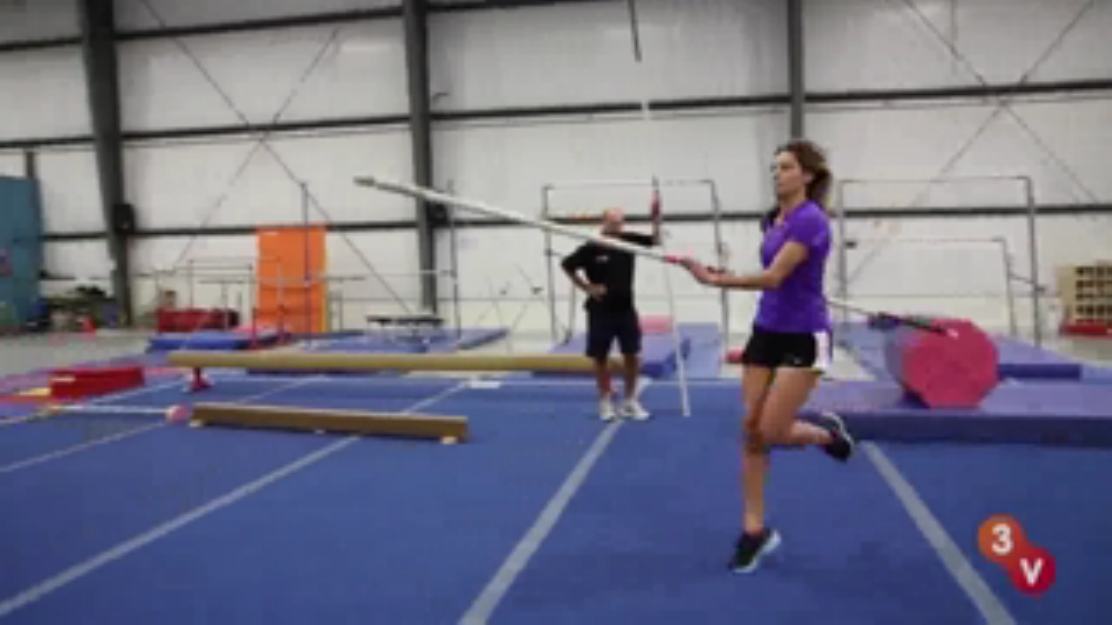}  \\
\includegraphics[width=0.98\linewidth]{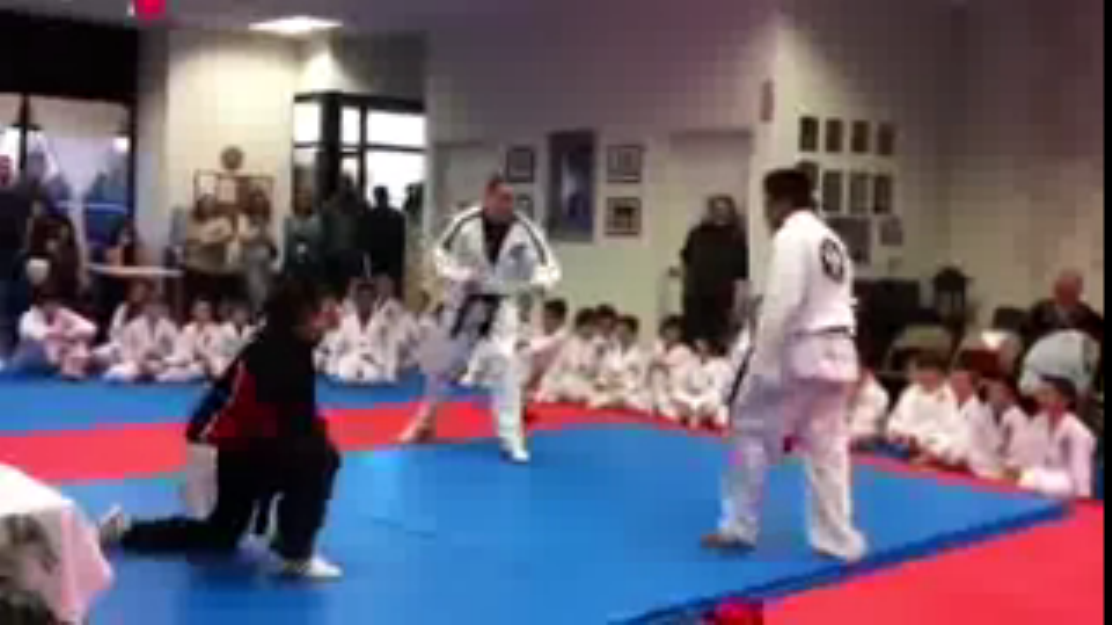}
\end{minipage}
\hfill
\begin{minipage}{0.19\linewidth}
\centering  {\bf Smelling food}
\includegraphics[width=0.98\linewidth]{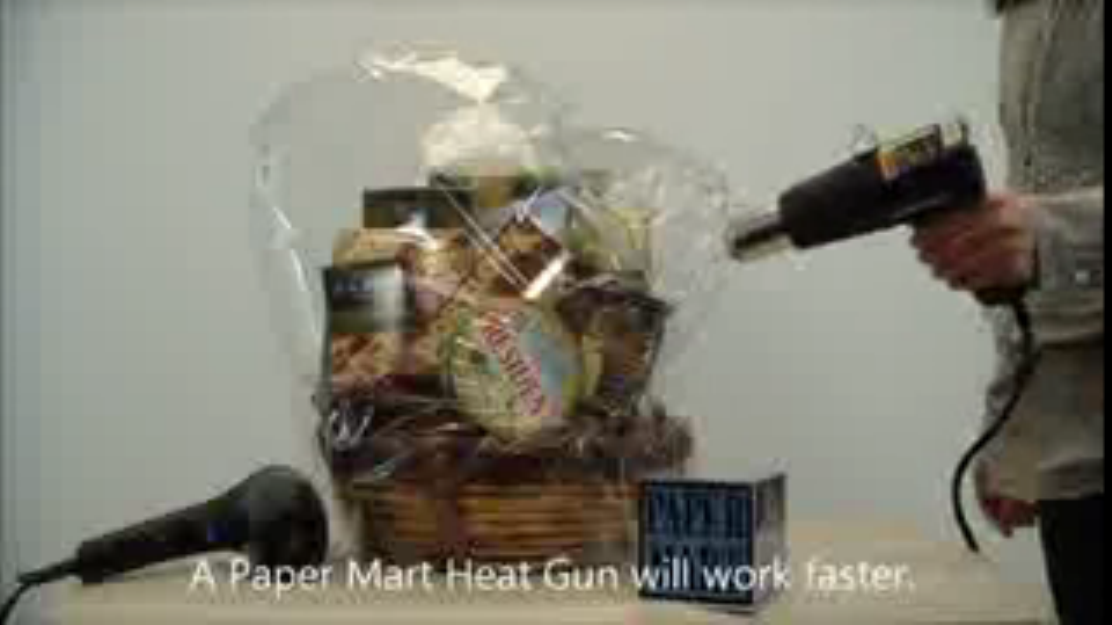}          \\
\includegraphics[width=0.98\linewidth]{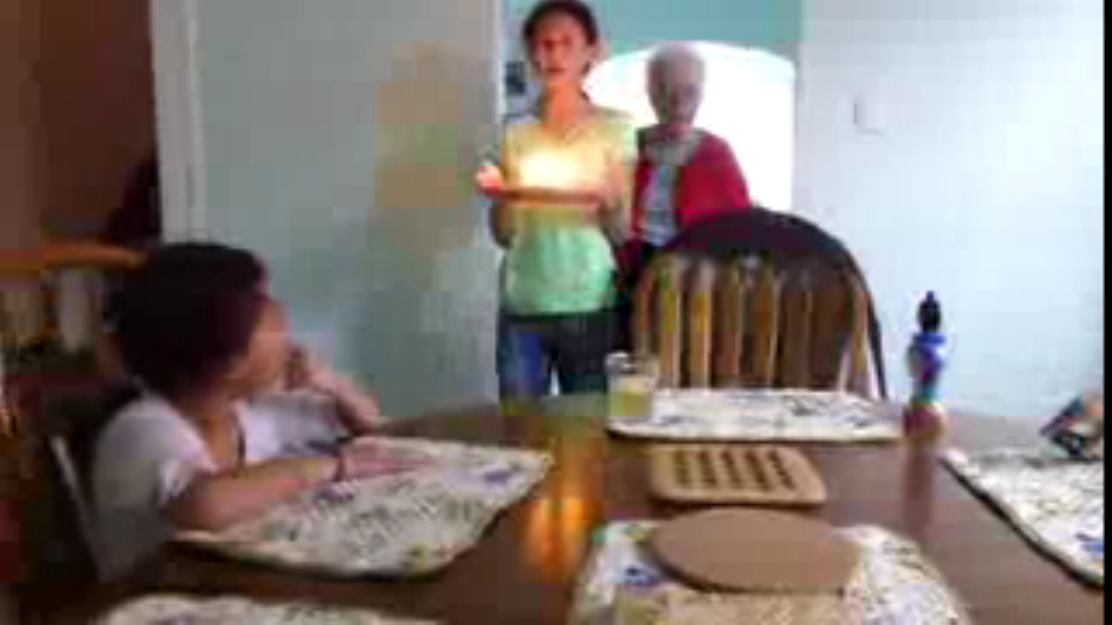}          \\
\includegraphics[width=0.98\linewidth]{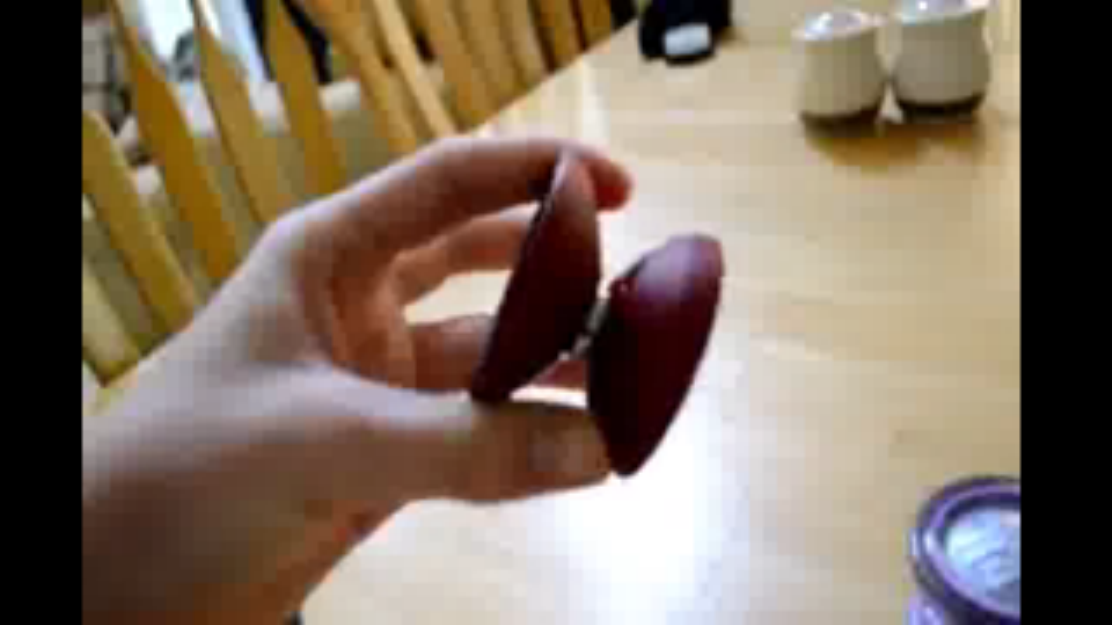}          \\
\includegraphics[width=0.98\linewidth]{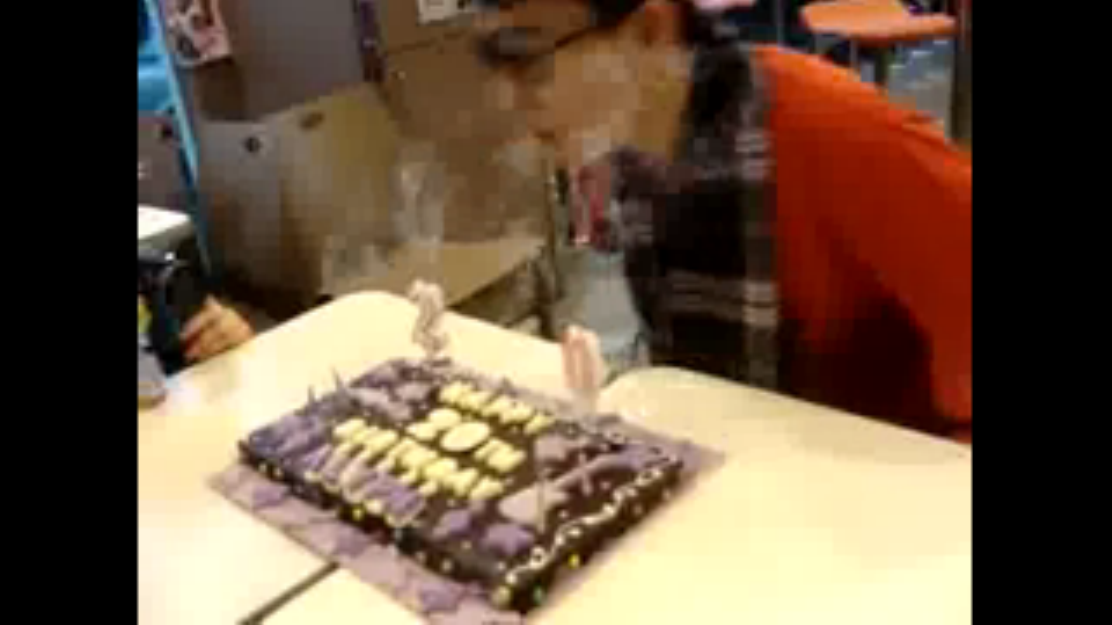}          \\
\includegraphics[width=0.98\linewidth]{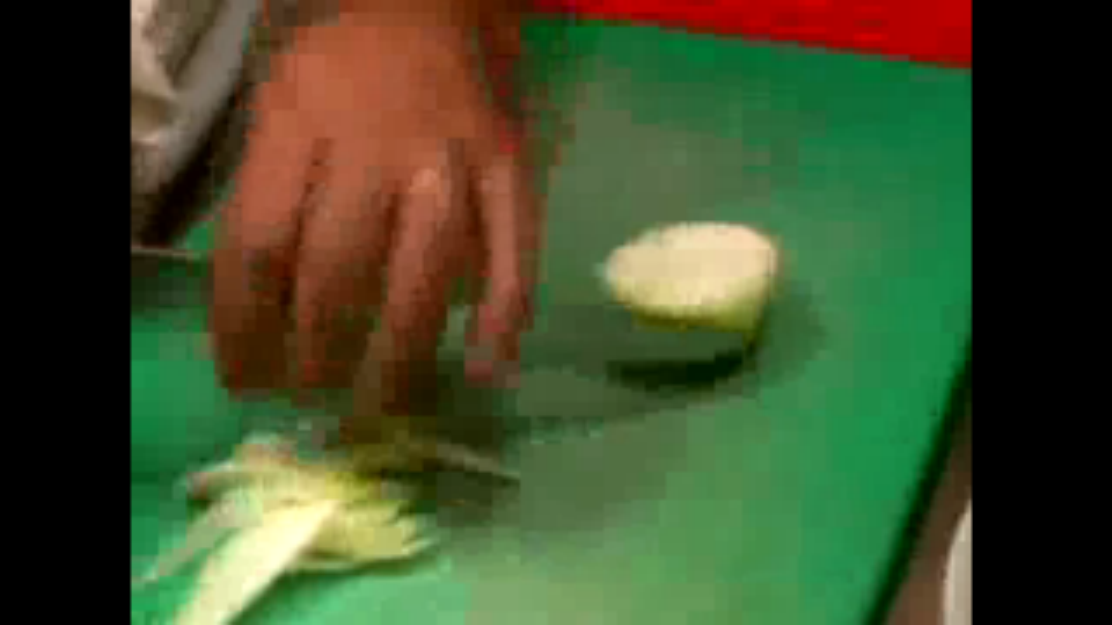}
\end{minipage}
\hfill
\begin{minipage}{0.19\linewidth}
\centering  {\bf Hit wicket}
\includegraphics[width=0.98\linewidth]{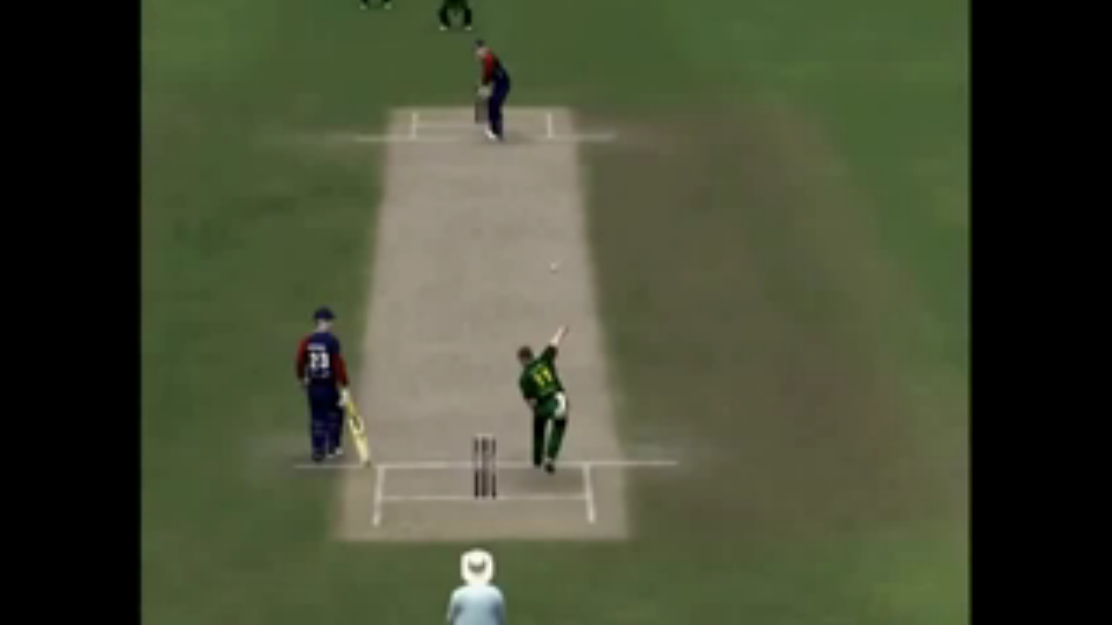}          \\
\includegraphics[width=0.98\linewidth]{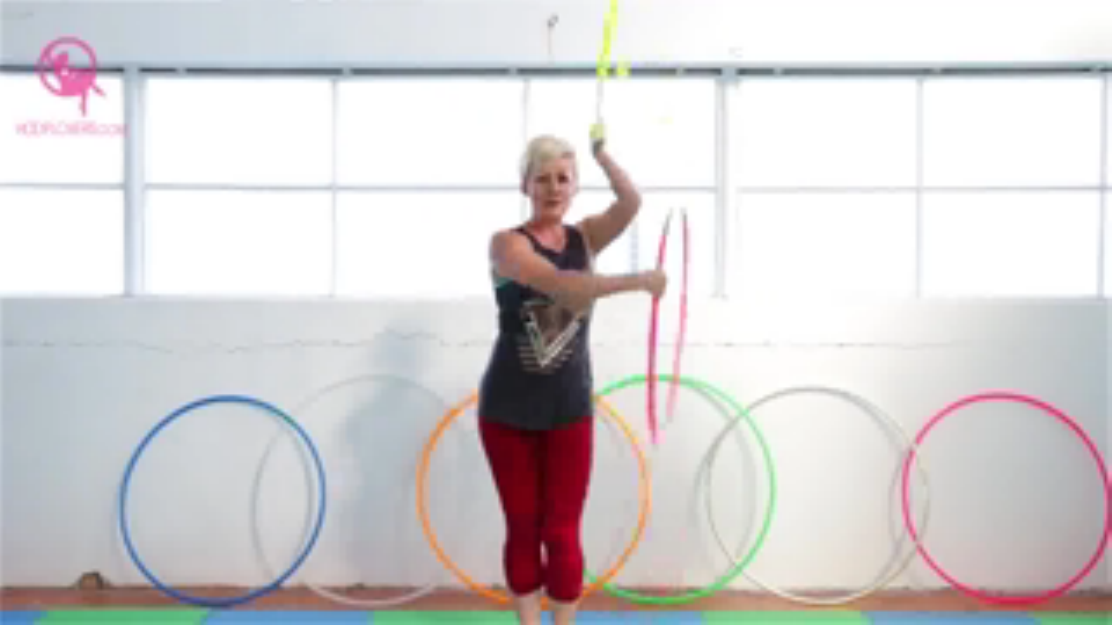}          \\
\includegraphics[width=0.98\linewidth]{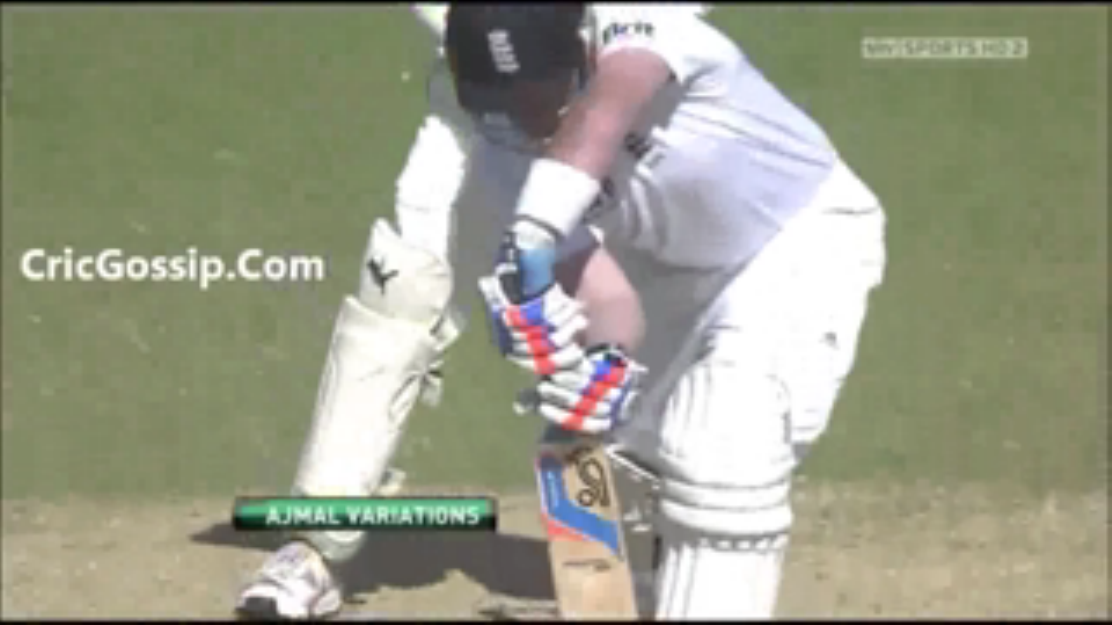}          \\
\includegraphics[width=0.98\linewidth]{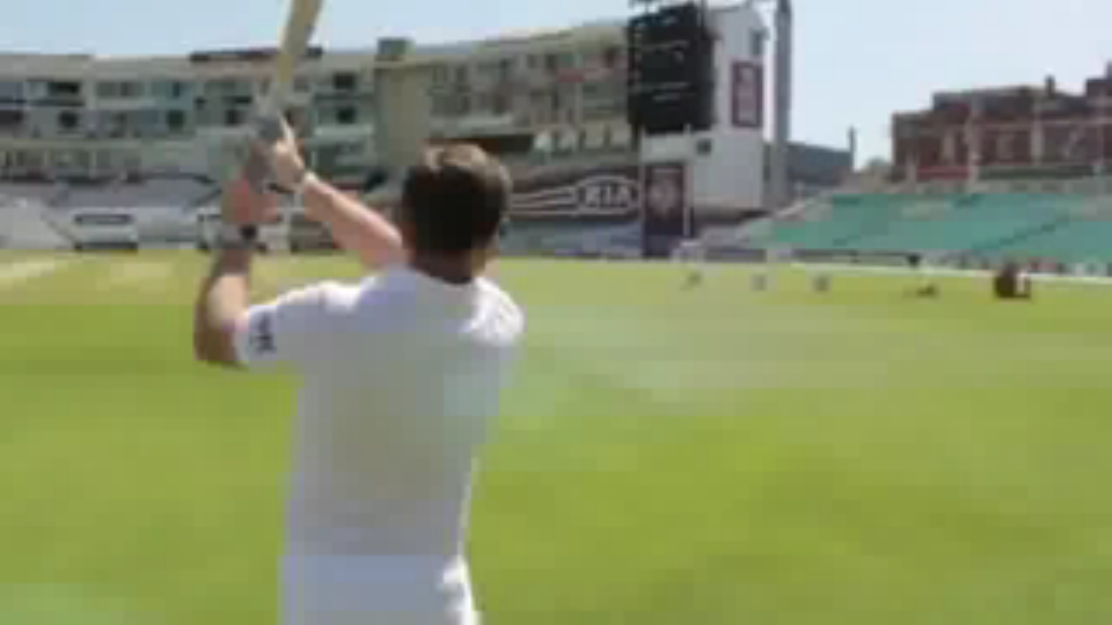}          \\
\includegraphics[width=0.98\linewidth]{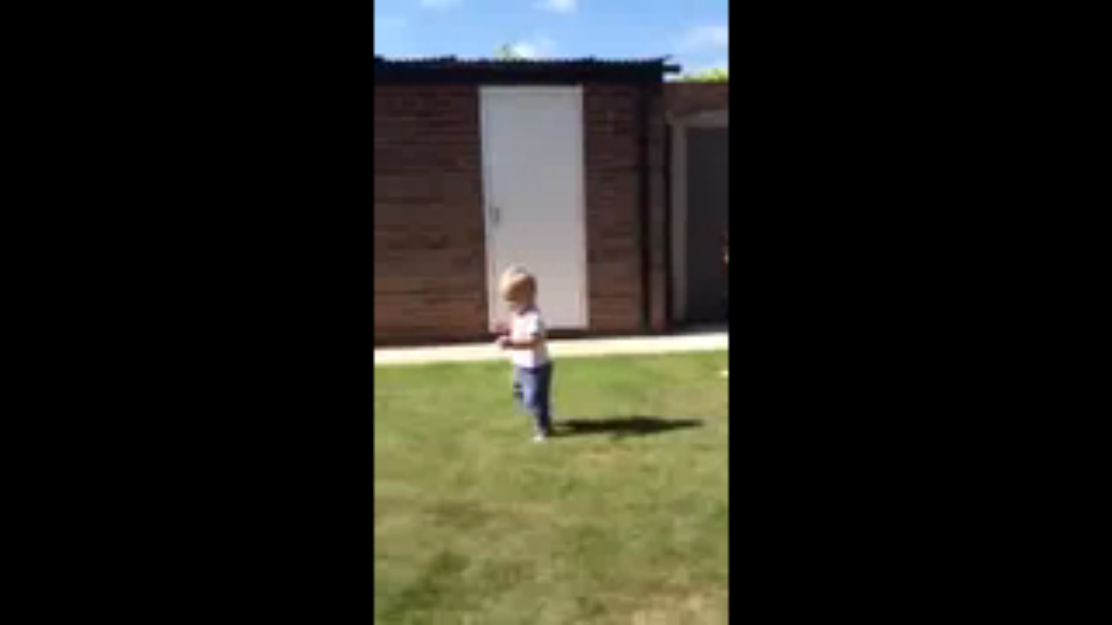}
\end{minipage}
~ \smallskip \\
\end{minipage}

\caption{Illustration of never seen actions on THUMOS14 testset: For a given textual action query the top five retrieved videos are shown.
The 101 classes of THUMOS14 do not contain these five action label queries. The first two queries are somewhat close to classes `\emph{Sumo wrestling}' and `\emph{Salsa spin}'. All retrieved videos for the query: `Fight in ring' include sumo wrestling. The videos retrieved for the second query: `Dancing' also includes two instances of dancing other than salsa. All results for the these two queries are technically correct. The third query `Martial arts' finds  mostly gymnasts, and a karate match. The fourth query is: `Smelling food', where we still obtain cakes, food items and dining table in the background. For the fifth query: `hit wicket' (in cricket) we do not succeed but retrieve some cricket videos. This illustration shows the potential for free keyword querying of action classes without using any examples.}
\label{fig:visualExamples}
\end{figure*}


In our final experiment, we aim to localize actions in videos, \ie, detect when and where an action of interest occurs.
We evaluate on the UCF Sports dataset, following the latest convention to localize an action spatio-temporally as a sequence of bounding boxes~\cite{Jain:tubelets,tian_iccv11,Yicong:sdpm}.
For sampling the action proposal, we use the tubelets from~\cite{Jain:tubelets} and compute object responses for each tubelet of a given video. 
We compare with the fully supervised localization using the object and motion representations described in Section~\ref{sec:act_cls}.
The top five detections are considered for each video after non-maximum suppression. 

The three are compared in Figure~\ref{fig:ST_loc}, which plots area under the ROC (AUC) for varying overlap thresholds. 
We also show the results of another supervised method of Lan \etal~\cite{tian_iccv11}. It is interesting to see that for higher thresholds our approach performs better. Considering that we 
do not use any training example it is an encouraging result.
There are other state-of-the-art methods~\cite{Jain_15kObjAct,Jain:tubelets,Yicong:sdpm} not shown in the figure to avoid clutter. These methods achieve performance comparable to or lesser than
our supervised case.

For certain action classes many objects and scene from the context might not be present in the groundtruth tubelets. Still our approach finds enough object classes for recognizing the zero-shot classes in the tubelets, as we have large number of train classes. 
In contrast, finding atomic parts of actions such as `look-up', `sit-down', `lift-leg' etc are difficult to collect or annotate.
This is one of the most critical advantages we have with objects, that it is easier to find many object or scene categories.

\section{Conclusion} 
We presented a method for zero shot action recognition without using any video examples. Expensive video annotations are completely avoided by using abundantly available object images and labels and a freely available text corpus to relate actions into an object embedding. In addition, we showed that modeling a distribution over embedded words with the Fisher Vector is beneficial to obtain a more precise sense of the unseen action class topic, as compared to a word embedding based on simple averaging. We explored sparsity both in the object embedding, as well as in the unseen action class, showing that sparsity is beneficial over mere feature-dimensionality. 

{We validate our approach on four action datasets and achieve promising results for action classification and localization. We also demonstrate our approach for action and event retrieval on THUMOS14 and TRECVID13 MED respectively.}
The most surprising aspect of our objects2action is that it can potentially find any action in video, without ever having seen the action before.

\vspace{2mm}
{
\small
\textbf{Acknowledgments}
This research is supported by the STW STORY project and the Dutch national program COMMIT.
}

{\footnotesize
\bibliographystyle{ieee}
\bibliography{egbib,tm}
}

\end{document}